\documentclass[letterpaper]{article} 
\usepackage[submission]{aaai23}  
\usepackage{times}  
\usepackage{helvet}  
\usepackage{courier}  
\usepackage[hyphens]{url}  
\usepackage{graphicx} 
\usepackage{natbib}  
\usepackage{caption} 
\usepackage{algorithm}
\usepackage{algorithmic}

\usepackage{comment}
\usepackage{todonotes}
\usepackage{subcaption}

\newcommand{\ie}{{\it i.e.},~}

\usepackage{newfloat}
\usepackage{listings}
\DeclareCaptionStyle{ruled}{labelfont=normalfont,labelsep=colon,strut=off} 
\floatstyle{ruled}
\newfloat{listing}{tb}{lst}{}
\floatname{listing}{Listing}
\title{The Value of AI Guidance in Human Examination\\ of Synthetically-Generated Faces}
\author{
    Aidan Boyd,
    Patrick Tinsley,
    Kevin Bowyer,
    Adam Czajka
}
\affiliations{
    University of Notre Dame\\
    Notre Dame, Indiana 46556, USA\\
    \{aboyd3,ptinsley,kwb,aczajka\}@nd.edu

}

\begin{document}

\maketitle

\begin{abstract}

Face image synthesis has progressed beyond the point at which humans can effectively distinguish authentic faces from synthetically generated ones. Recently developed synthetic face image detectors boast ``better-than-human'' discriminative ability, especially those guided by human perceptual intelligence during the model's training process. In this paper, we investigate whether these human-guided synthetic face detectors can assist non-expert human operators in the task of synthetic image detection when compared to models trained without human-guidance. We conducted a large-scale experiment with more than 1,560 subjects classifying whether an image shows an authentic or synthetically-generated face, and annotating regions supporting their decisions. In total, 56,015 annotations across 3,780 unique face images were collected. All subjects first examined samples without any AI support, followed by samples given (a) the AI's decision (``synthetic'' or ``authentic''), (b) class activation maps illustrating where the model deems salient for its decision, 
or (c) both the AI's decision and AI's saliency map. Synthetic faces were generated with six modern Generative Adversarial Networks. Interesting observations from this experiment include: (1) models trained with human-guidance offer better support to human examination of face images when compared to models trained traditionally using cross-entropy loss, (2) binary decisions presented to humans results in their better performance than when saliency maps are presented, (3) understanding the AI's accuracy helps humans to increase trust in a given model and thus increase their overall accuracy. This work demonstrates that although humans supported by machines achieve better-than-random accuracy of synthetic face detection, the approaches of supplying humans with AI support and of building trust are key factors determining high effectiveness of the human-AI tandem.

\end{abstract}

\section{Introduction}

\begin{figure}[!htb]
        \centering
            \includegraphics[width=\columnwidth]{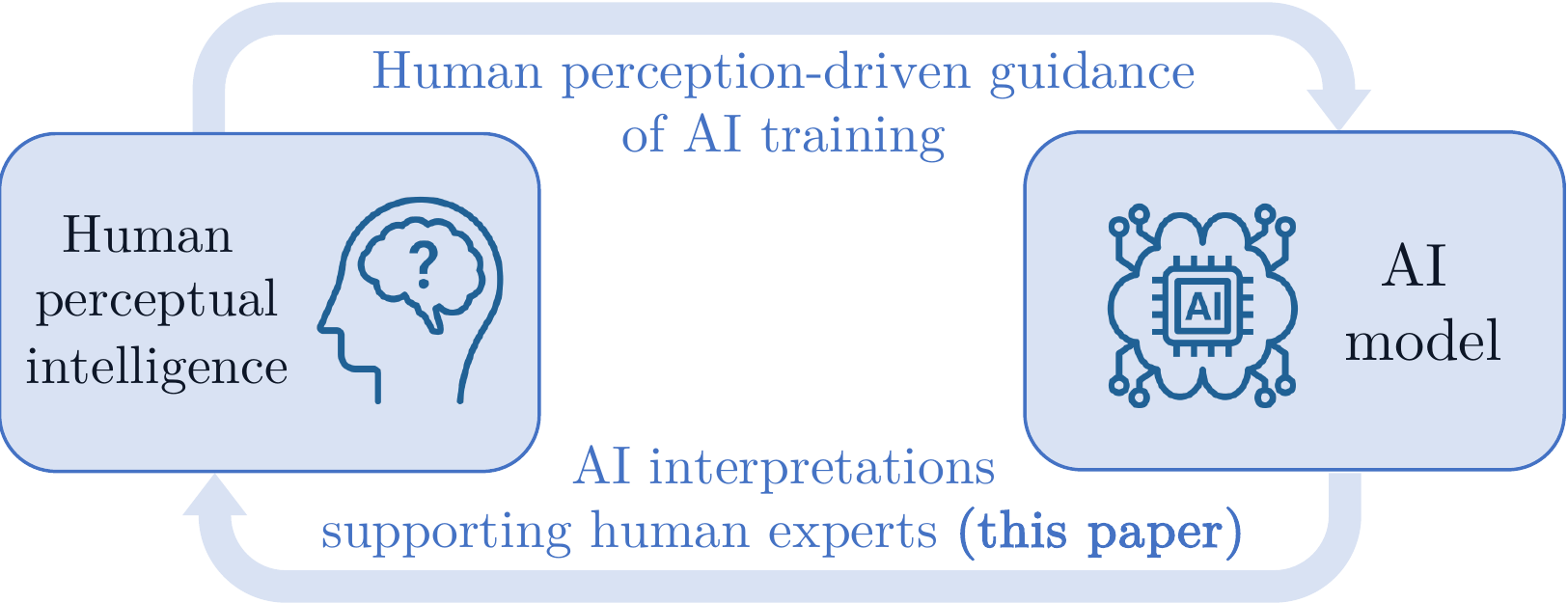}
            \captionof{figure}{The human-AI supervision cycle, in which training of the AI is first guided by human perceptual intelligence to increase AI's generalization capabilities, and then the AI supervises humans examining unknown samples. We explore this approach in the context of synthetic face detection and, focusing on the lower part of the cycle,  investigate the value added by the AI when supporting humans in detecting synthetic images.}
            \label{fig:teaser}
\end{figure}

Visual realism of synthetically-generated faces is now so high that humans, for whom face recognition is a core social capability, cannot effectively distinguish authentic and synthetic face images \cite{shen2021study}. Recent studies even suggest that face images synthesized by modern generative models may be perceived by humans as more ``trustworthy'' than authentic faces \cite{doi:10.1073/pnas.2120481119}. There are, however, situations, in which reliable judgment about face image authenticity is required by both laymen and professional examiners. The former relates to, for instance, our capabilities to safely exist in social media environment, and the latter relate to forensic examination of face images.

A natural next step toward improving human capabilities in detecting synthetic faces is to pair AI and humans together into a supervision cycle (Fig. \ref{fig:teaser}). In this ``partnership'', both humans and AI guide their counterparts to salient information, helping them to understand the problem and ultimately make better decisions. In \cite{boyd2021human,boyd2021cyborg}, the first step of the human-AI supervision cycle (humans guiding the AI) was implemented for the task of synthetic iris and face detection. Human annotations of salient image regions were either used to augment the training data, or were incorporated into the loss function to guide the network's training process, resulting in better generalization against unseen image types. Taking inspiration from those works, this paper proposes the first known attempt at implementing the second step of the human-AI supervision cycle (AI guiding human subjects) for the task of synthetic face detection. 

More specifically, we summarize results of the large-scale experiment of synthetic face detection carried out by 1,560 humans supported by deep learning-based synthetic face detection algorithms, trained with and without prior human guidance, and offering explanations of their decisions in the form of nowadays-popular class activation maps (CAM). All subjects were asked to annotate features supporting their decision in a specially-designed tool, and 56,015 annotations across 3,780 unique face images were collected. As a control experiment, all subjects first classified and annotated 18 samples without any AI support. Then, 18 new face images were presented along with one of: the AI's decision, the AI's salience (CAM), or both decision and salience.

We make a few immediate observations from this study. First, we observe that a popular explainability technique using CAM to highlight the model's salient regions {\it does not} help humans to improve their accuracy, whereas presenting only the AI's decision {\it does} lead to a better accuracy of human-AI tandem, compared to humans working alone. Second, models trained with human-guidance offer {\it better} human support (measured as increased performance of human-AI pair). Third, informing humans about the AI's accuracy {\it does} lead to a better agreement between the algorithms and the subjects, suggesting that people have more trust in AI if they know its even basic performance numbers. Owing to the abundant and multimodal data collected for this study, we define a series of detailed {\bf research questions} and organize the paper in a way to provide answers to these questions:

    \vskip1mm\noindent
    {\bf -- RQ1:} How do non-expert subjects perform in the task of detecting synthetic faces generated by modern Generative Adversarial Networks when they are explicitly instructed to annotate salient features?
    
    \vskip1mm\noindent
    {\bf -- RQ2:} Are some of the image sources (generative models) more difficult for humans to classify than others?
    
    \vskip1mm\noindent
    {\bf -- RQ3:} Does time spent on examining an image, or the annotation area, affect human accuracy? 
        
    \vskip1mm\noindent
    {\bf -- RQ4:} What is the difference in accuracy between the scenario with no AI support and the various types of auxiliary AI information? Do some types of AI support help the subjects more than other types? 

    \noindent
    {\bf -- RQ5:} What is the agreement between AI and humans? How does this agreement change depending on the information offered along with images (decision / salience)? Do humans agree with human-guided models more than traditionally trained models?

    \vskip1mm\noindent
    {\bf -- RQ6:} Do humans implicitly recognize an AI's true accuracy from observing its decisions without being informed?  
    
    \vskip1mm\noindent
    {\bf -- RQ7:} Does adding this explicit knowledge of the AI's accuracy translate to higher or lower accuracy and agreement between humans and AI? Does informing people about high AI accuracy build trust in AI? Does it make people over-trust the AI? 

This paper describes the first known experiment of human examination of synthetic and authentic faces in which subjects are supported by two types of AI (trained with and without human salience) in three different ways (presentation of decision, model's salience, or both). We will also release the entire dataset of 56,015 annotations and reaction times, which can facilitate future research on human-AI pairing in the context of synthetic face detection with the published version of this paper.

\section{Related Work}

\paragraph{Human-AI Pairing}

The recent renaissance of humans-machine teaming  is caused by spectacular advances in deep learning, and an abundant number of tools that allow for experimenting with neural network learning and architecture design. 
This subsection focuses on vision and human perception intelligence, in which we see both disappointing and promising results of putting humans and machines into collaboration. Starting from the former, \cite{Jacobs_Nature_2021} presented patient vignettes to medical doctors, each vignette with or without the AI's recommendation and a form of explanation, and found that ``interacting with machine learning recommendations {\it did not} significantly improve clinicians’ treatment selection accuracy.'' \cite{Bansal_CHI_2021} paired AI and humans to solve visual classification tasks, and {\it did not} find a significant improvement from explanations offered by an AI. Even worse: ``explanations increased the chance that humans will accept the AI’s recommendation, regardless of its correctness.''. Conversely, combining human and algorithm capabilities in the biometric context had positive outcomes. \cite{OToole_TSMC_2007} demonstrated that fusing humans and algorithms {\it increased} face recognition accuracy to near-perfect values on the Face Recognition Grand Challenge dataset. \cite{He_ICCV_2019} concluded that human and algorithm visual saliencies differ and their integration {\it increases} the quality of image captioning. \cite{Moreira_WACV_2019} studied how recognition of challenging post-mortem irises differs between humans and machines, and demonstrated that fusing results from humans and algorithms end up with a more accurate method than either alone \cite{Trokielewicz_BTAS_2019}. 

\begin{figure*}[!ht]
    \centering
    \includegraphics[width=0.7\linewidth]{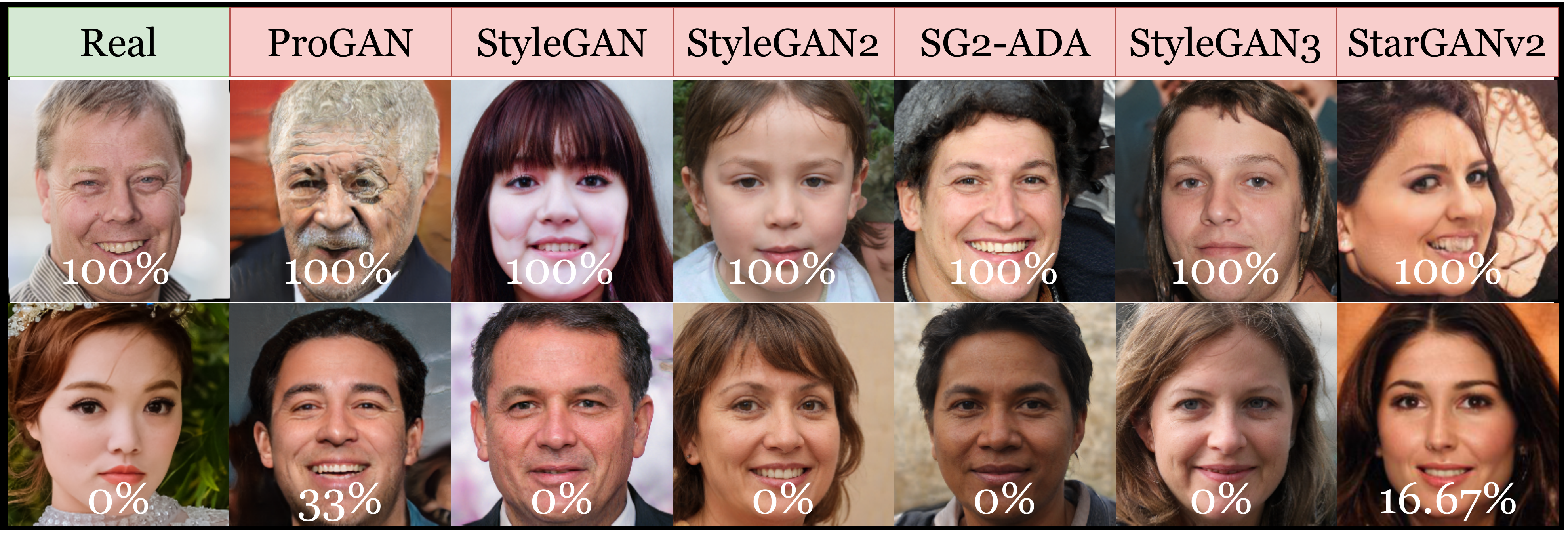}
    \caption{Examples from each data source. Top row: One sample from each of the seven image sources (one authentic plus six synthetic) that was the easiest for human subjects to correctly classify in the control portion of the collection. Bottom row: the most difficult samples to correctly classify by human subjects. The percentages shown correspond to the average human accuracy on that image. Each of these images were classified by six independent subjects.
    }
    \label{fig:test_data}
\end{figure*}

\paragraph{Synthetic Image Generation}

Open-source tools for generating synthetic face images 
generally use Generative Adversarial Networks~\cite{goodfellow2014generative}. 
Early synthetic face generation was accomplished by progressive GAN \cite{karras2017progressive}, which further evolved into models disentangling various attributes of faces, such as hairstyle \cite{karras2019style}, improving on generation-specific artifacts \cite{karras2020analyzing,Karras2021}, and addressing limited training data \cite{Karras2020ada,brock2018large}. These {\it StyleGAN} architectures became widely known due to the popularity of the \url{thispersondoesnotexist.com} webpage.
Other approaches create new identities by transferring the  style of a source face \cite{choi2020stargan} to a new identity. The state-of-the-art models are now able to synthesize visual samples that are often indistinguishable from authentic ones by humans.

\paragraph{Human Trust in AI}

Human-machine teaming, rather than simply merging/fusing human and machine decisions, is a promising path to use AI for good. However, this will not succeed without building trust into the AI's design by making it not only more reliable, but also more transparent and explainable. Both over-trust in limited  AI, and under-trust in reliable AI, generate risks related to lower reliability of human-machine teams, compared to each component alone, as discussed in \cite{wong2020much}. 

Human trust in AI is a multidimensional and difficult-to-measure issue.
In the biometrics context, trust spans across several dimensions such as performance, fairness, security, explainability and interpretability, and privacy \cite{Jain_TBIOM_2022}. Pairing of biometric systems with humans complicates this issue as, in addition to placing trust in a system, we need to add aspects of human trust into the examined data and its source. 
\cite{doi:10.1073/pnas.2120481119} demonstrated that synthetically-generated faces are nearly indistinguishable from authentic faces and are judged by humans as more trustworthy. However, mere awareness of the existence of synthetic (GAN-generated) in public domain may erode human trust \cite{Tucciarelli_PsyArXiv_2020}. If additionally we show authentic faces to humans along with a wrong classification of being synthetic, such samples may be judged to be less trustworthy than the same images without such wrong label \cite{Liefooghe_PsyArXiv_2022}. \cite{Nakano_HSSC_2022} showed that we trust more in self-resembling faces than those judged by a neural network-based face recognizer as more distant. These examples demonstrate limited human capabilities to spot synthetic faces, reinforcing that appropriate AI support may be crucial to solve such tasks.

\section{Data Collection}

\begin{figure*}[!htb]
    \centering
    \includegraphics[height=0.36\columnwidth]{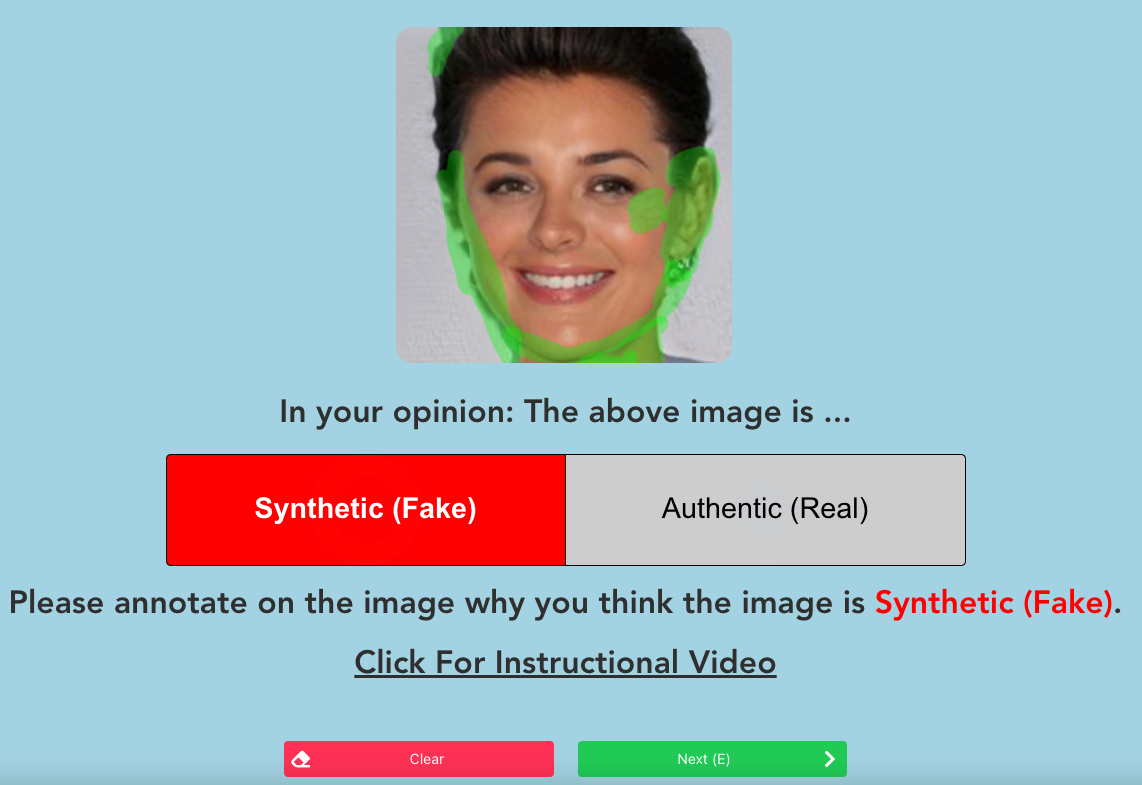}\hfill
    \includegraphics[height=0.36\columnwidth]{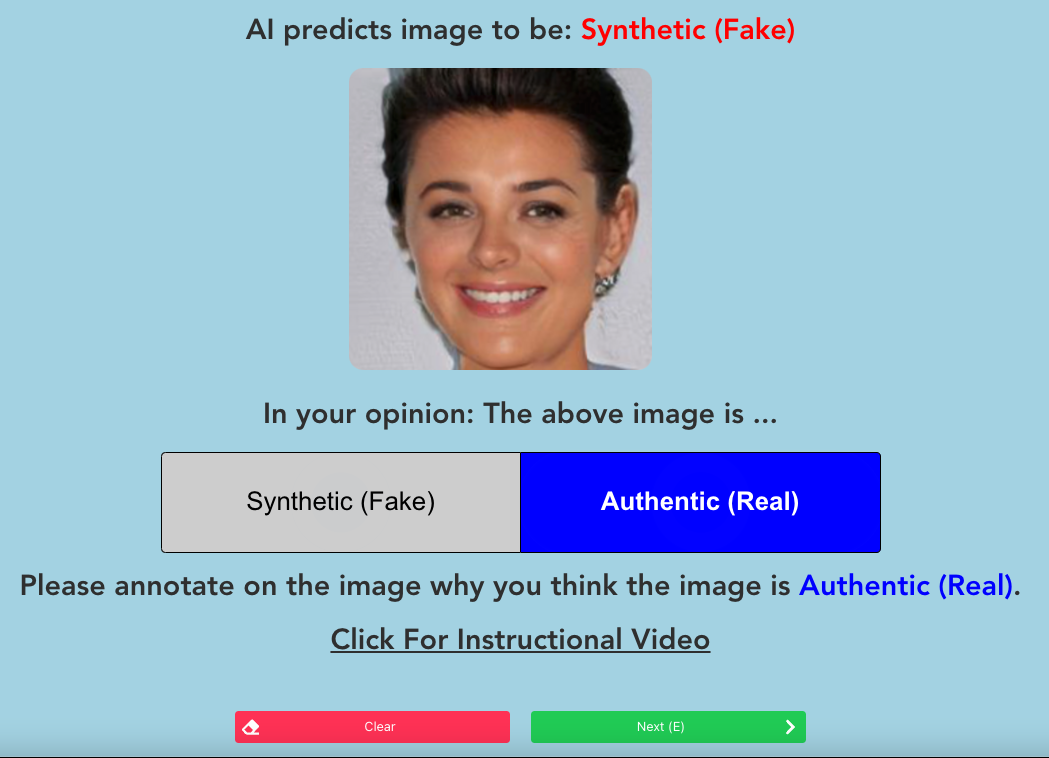}\hfill
    \includegraphics[height=0.36\columnwidth]{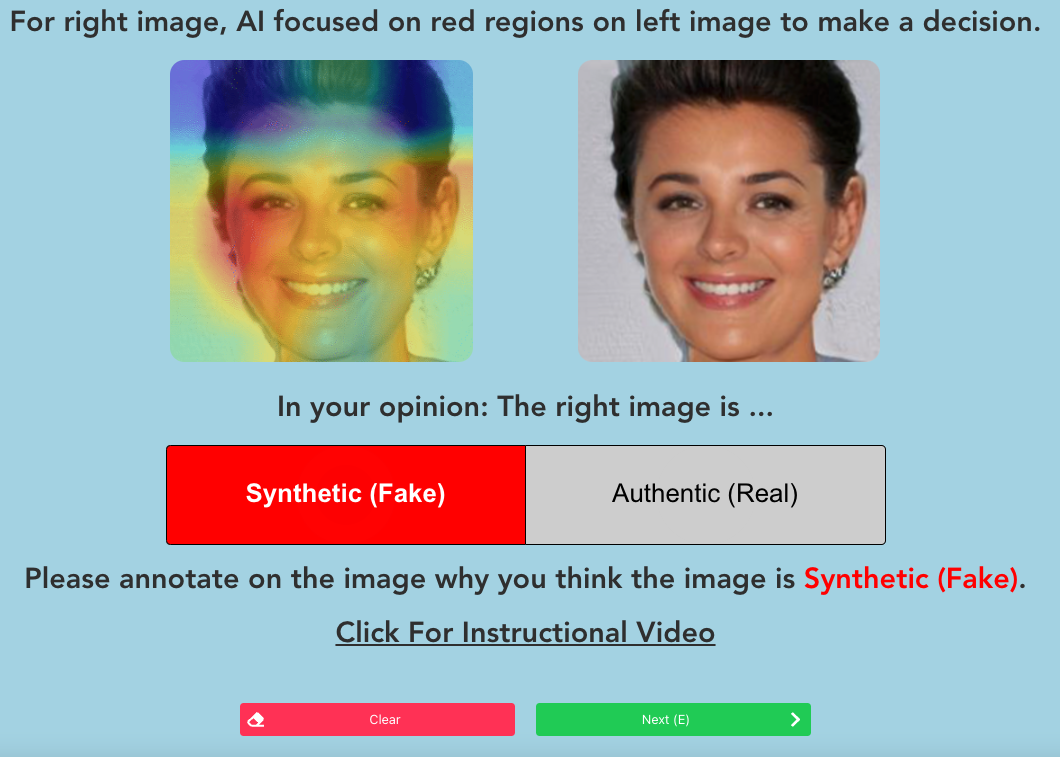}\hfill
    \includegraphics[height=0.36\columnwidth]{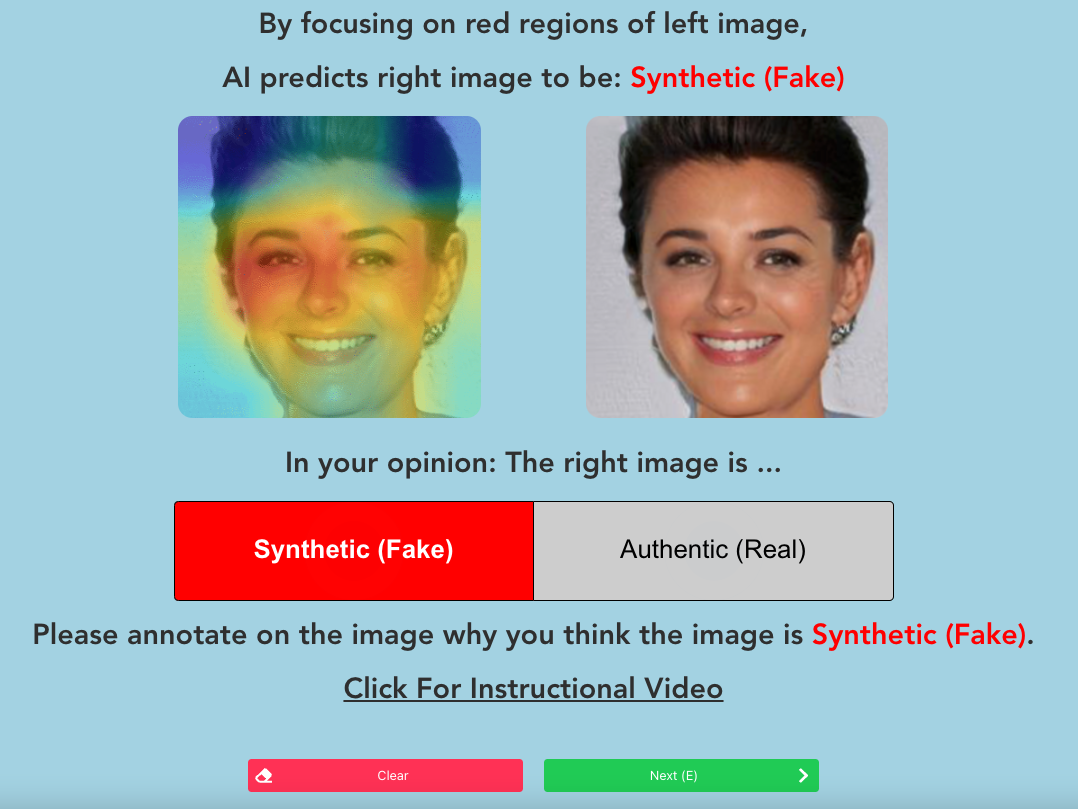}
    \caption{The online tool used in the experiments and three types of AI support offered to humans ({\bf from left to right:}) no AI support (with example annotations), only AI's classification (authentic/synthetic) of the face image displayed, only AI's saliency (CAM) overlayed on the examined face image,  and both the AI's saliency and decision showed to humans.}
    \label{fig:annot_tool}
\end{figure*}

\subsection{Experimental Datasets}
\label{sec:datasets}

In this study, we use authentic images from Flickr-Faces-HQ~\cite{karras2019style}, and synthetic face images from six different generators (ProGAN, StyleGAN, StyleGAN2, StyleGAN2-ADA, StyleGAN3, and StarGANv2~\cite{karras2017progressive,karras2019style,karras2020analyzing,Karras2020ada,Karras2021,choi2020stargan}).
Fig. \ref{fig:test_data} and the following paragraphs briefly characterize data sources and preprocessing steps,

\vskip1mm\noindent{\bf Flickr-Faces-HQ (FFHQ)} includes 70,000 images of authentic faces varying in age, ethnicity, and facial accessories (glasses, hats, etc.) \cite{karras2019style}. 

\vskip1mm\noindent{\bf ProGAN} is the oldest GAN in this work, 
which was trained on CelebA-HQ images~\cite{karras2017progressive}.

\vskip1mm\noindent{\bf The StyleGAN Family} \cite{karras2019style,karras2020analyzing,Karras2020ada,Karras2021}. 
The original StyleGAN was trained in a similar manner to its predecessor ProGAN~\cite{karras2017progressive}, but with the added feature of mixable disentangled layers for style transfer.
StyleGAN2~\cite{karras2020analyzing} improved on this by removing irregular artifacts found in original StyleGAN images and improved image reconstruction via path length regularization.
The third iteration of StyleGAN, StyleGAN2 with adaptive discriminator augmentation)~\cite{Karras2020ada}, solves for training GANs in data-limited scenarios. 
Finally, StyleGAN3~\cite{Karras2021} mitigates aliasing in rotation- and translation-invariant generator networks.

\vskip1mm\noindent{\bf StarGANv2}, unlike StyleGAN, employs style transfer~\cite{choi2020stargan} via source and reference face images: source images are given the style of reference images to create new identities. 

\paragraph{Image Preprocessing.} All face images are aligned using \verb+img2pose+ \cite{Albiero_CVPR_2021}, cropped, and resized to 336×336. Face bounding boxes are expanded 20\% in all directions before cropping, with an additional 30\% on the forehead to ensure the entire head is in view.

\subsection{AI Cues Shown to Humans}
\label{sec:ai_descr}
To investigate whether AI models can help inform humans in a way to increase task accuracy, we can ask the AI model \textbf{what} is its decision and \textbf{why} it made that decision. Supplying the AI decision to humans is as simple as telling them whether the model believes the image to be authentic or synthetic. \textbf{Why} the model came to this decision, on the other hand, is more complex. The interpretation in this work of why the AI made that decision relates to areas in the image deemed most important for that decision. Thus, due to their simplicity,  
Class Activation Maps (CAMs) \cite{zhou2016learning} are employed to represent model saliency by creating a weighted sum of the activations in the network's final convolutional layer. Regions of the image that correspond to larger activations are given more weight in the final output.  

We use pre-trained DenseNet-121 models to detect synthetic faces \cite{boyd2021cyborg}
, and devise three separate experimental settings, as illustrated in Fig. \ref{fig:annot_tool}. The first, and most straightforward, is displaying the AI decision in sentence format (denoted as {\bf Decision Only}). The sentences used are \textit{AI predicts image to be: Synthetic (Fake)} or \textit{AI predicts image to be: Authentic (Real)}. This sentence is placed above the query image in clear view. Importantly, humans are unaware of the underlying accuracy of the AI models.

The second setting is displaying the AI's CAMs without any decision information (\ie telling humans where the AI ``looked'' but not what it decided, denoted as {\bf CAM Only}). 
CAMs are overlayed on the original image as a heatmap, where red and blue regions represent more and less salient regions, respectively. Because we could use 10 models in each training setting \cite{boyd2021cyborg}, we combine the generated CAM heatmaps into a GIF that smoothly transitions between all 10 unique CAMs over a 10s period. 

Finally, in the third setting, we show users both what the AI decided as well as where it looked to make that decision (denoted as {\bf Both Decision and CAM}). Using the same decisions and CAMs from the previous settings, this information is combined to give human subjects the most complete explanation of the AI's processing in this study. 

Additionally, this paper investigates the use of AI model types trained in two different ways for a task of synthetic face detection: (a) models trained traditionally with cross-entropy loss, and (b) human-guided models, trained in a way to focus more on human-salient features. 
Thus, for each of the three AI input settings presented above, experiments are completed for both {\bf traditionally trained models} and {\bf human-guided models}.

\subsection{Data Collection}

\paragraph{Acquisition Tool and Strategy.} Each image set shown to human annotators contains 36 unique samples and is broken into two phases: the control and the experiment. Both the control and the experiment are sets of 18 images, consisting of six authentic images, and two images from each of the six image synthesis sources detailed in Sec. \ref{sec:datasets}. The images are shown in random order, but this order is consistent across different annotators seeing that set of images. Image sets are set up such that each individual image will be part of the control for six different annotators and once for each of the six experimental settings detailed in \ref{sec:ai_descr}.

To facilitate the data collection, an online tool was developed. Subjects, recruited via Prolific platform (\url{https://www.prolific.co}), were presented an image of a face and asked to select whether they believe the nature of that image to be \textit{Authentic (Real)} or \textit{Synthetic (Fake)} in a two-alternative forced choice (2AFC) fashion, Fig. \ref{fig:annot_tool}. After one of those options are selected, users are then asked to draw, using their mouse, regions on the image that they believe led them to that decision. This annotation is referred to as human saliency \ie the regions deemed salient for the classification decision. Each image must be annotated before the user is permitted to move to the next sample. A ten second minimum time was imposed on each image to encourage annotators to spend more time and care on each sample. Prolific workers were paid \$4.50 and the average completion time was 23.8 minutes. This was in line with the recommended hourly rate specified by Prolific. 

\paragraph{Quality Checks During and After Experiment.} Due to the inherent noise in online data collections, two quality checks were instituted. The first included a detailed video demonstrating the operation of the tool and correct annotation of salient regions. Subjects had to watch the video and provide a code embedded into it to start the experiment.

The second quality check involved the manual examination of the collected annotations. Any sets of annotations from individual users that blatantly misunderstood the task or were clearly \textit{speed-running}, were discarded and that image set was put back online for another random user to complete. To make sure no bias was introduced as to what represents a ``good annotation'' in this manual checking step, the authors were very lenient in what constituted an acceptable annotation set; as long as users showed they were completing the task at hand, their experiment was accepted.

Data was collected from $1,260$ unique human annotators totaling $45,232$ individual, good-quality annotations in this set of experiments. This represents multiple annotations from 12 different human examiners on 3,780 images (1,260 images of authentic faces and 2,520 images of synthetic faces). Further experiments included 300 more subjects and $10,833$ additional annotations totalling 1,560 subjects and 56,065 annotations. 

\section{Experimental Results}

\begin{figure}
    \centering
    \includegraphics[width=\linewidth]{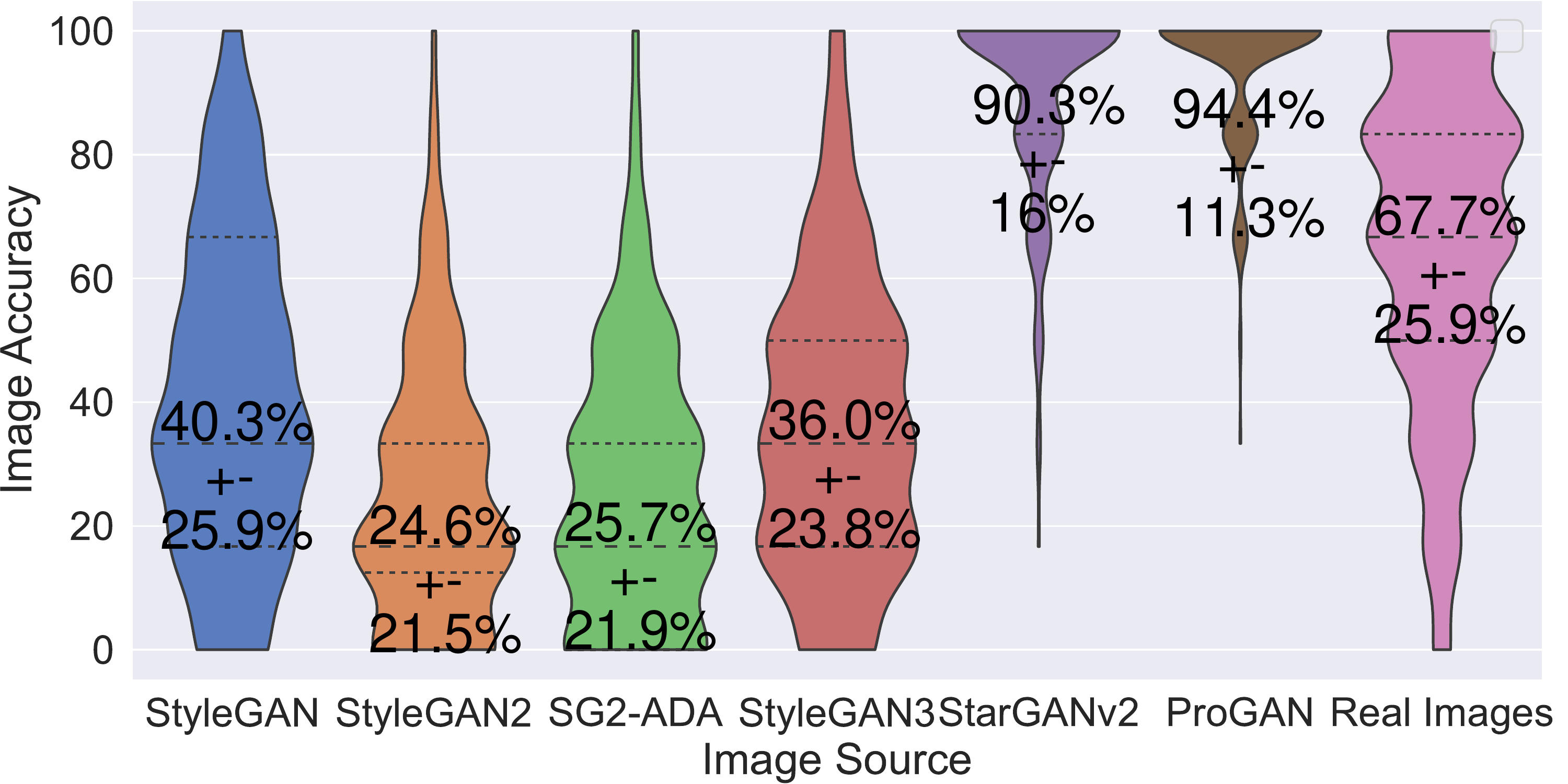}
    \caption{Accuracy of human subjects for synthetic and authentic face images originating from different sources without AI support. Values represent mean image accuracy $\pm1\sigma$.}
    \label{fig:gan_acc}
    \null
\end{figure}

This section describes the results,
organized around  questions posed in the Introduction.  
Mean accuracy is used as the main metric, presented with $\pm1\sigma$ across the subjects. 

\paragraph{Answering RQ1 (Novice accuracy in detecting synthetic faces).}

The no-AI-input (or ``control'') phase of the task consisted of six authentic faces and twelve synthetic faces.
This phase establishes baseline human accuracy at synthetic face detection. Authentic face images were correctly classified $65.7\%\pm23.63\%$ of the time and synthetic images $51.88\%\pm15.51\%$. 
These results show that {\bf novice humans solve  synthetic face detection at better-than-random accuracy} in our scenario.

\paragraph{Answering RQ2 (Human accuracy on synthetic faces from different sources).}

Fig. \ref{fig:gan_acc} illustrates the \textit{control} accuracy across the image sources. 
We can see that ProGAN and StarGANv2 are the most easily detected of the synthetic images. Next easiest to classify are authentic images. Both StyleGAN and StyleGAN3 show slightly below random chance human accuracy. The most difficult to classify are StyleGAN2 and its update StyleGAN2-ADA (SG2-ADA on figure), each classified correctly only about 1 out of every 4 images. Humans on average classified samples from StyleGAN2 and StyleGAN2-ADA as authentic more than the actual authentic images. Thus, from Fig. \ref{fig:gan_acc} it can be concluded that {\bf there is a wide range of accuracy across data sources}, with StyleGAN2 and StyleGAN2-ADA being the most difficult.

\paragraph{Answering RQ3 (Human accuracy varying with examination time).}

\begin{figure}
    \centering
          \includegraphics[width=0.48\columnwidth]{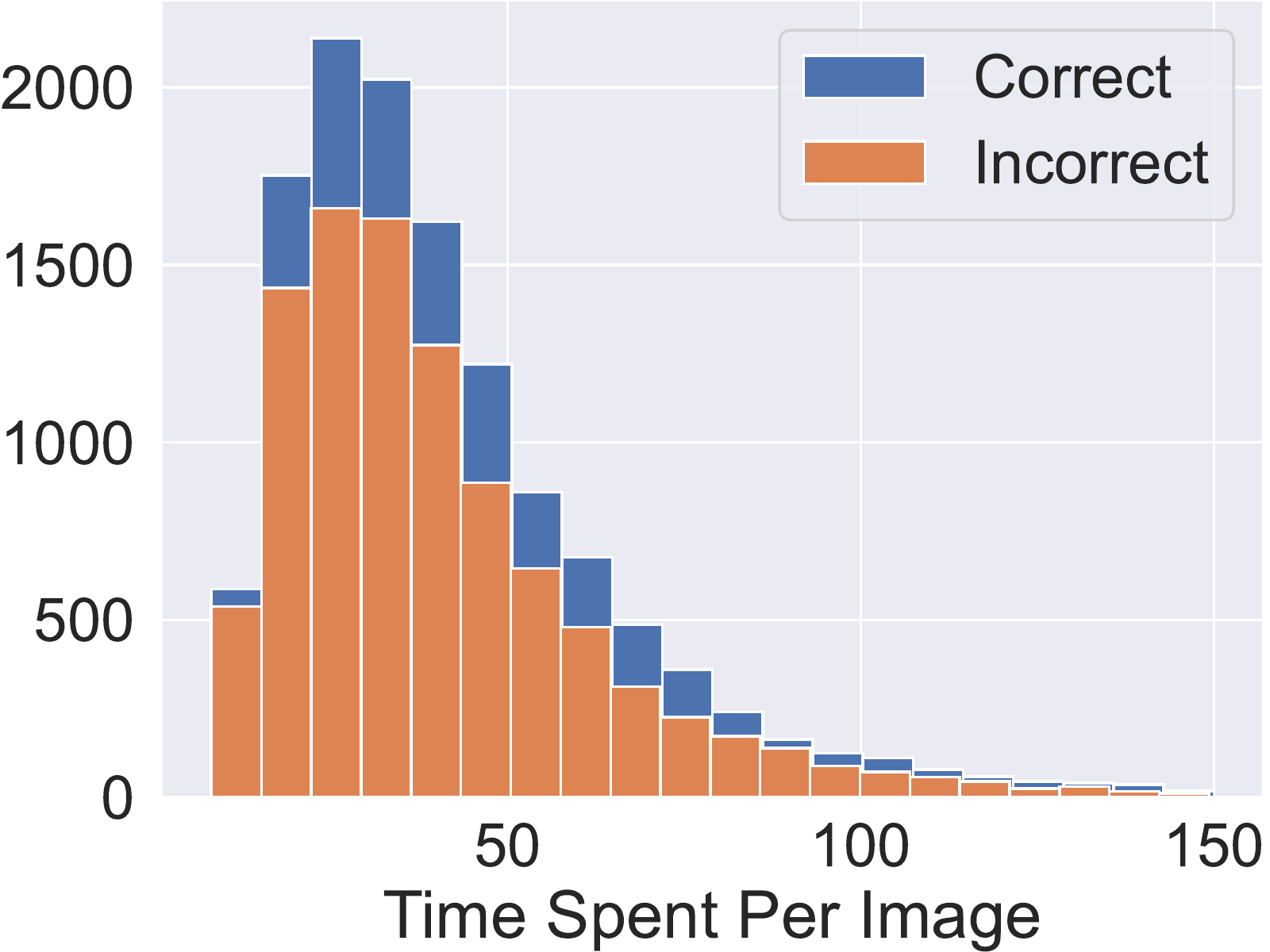}\hfill
          \includegraphics[width=0.48\columnwidth]{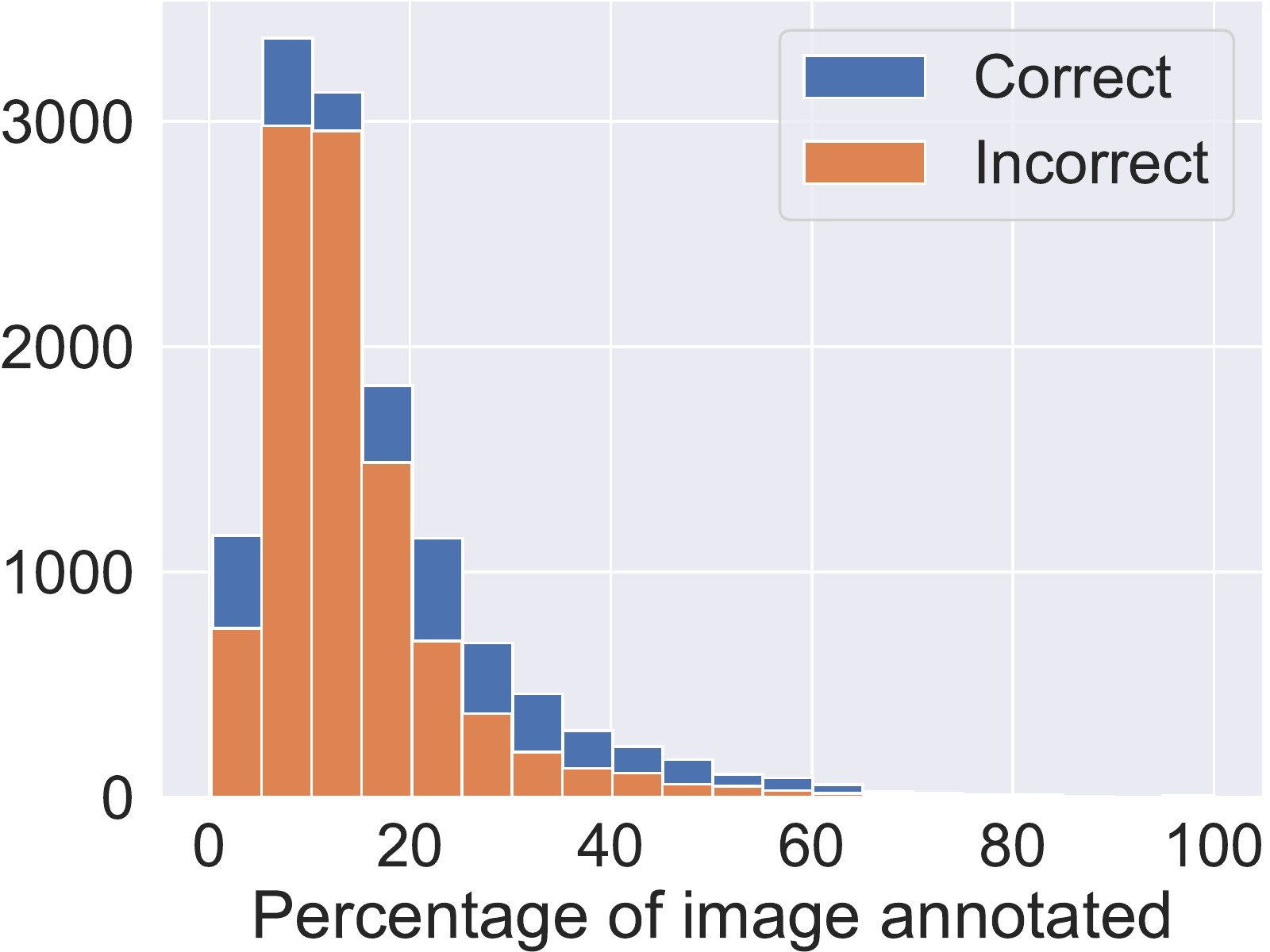}
    \caption{Time in seconds ({\bf left}) and annotation area ({\bf right}) for correct and incorrect (overlayed) human classifications. Subjects had to spend minimum 10 seconds on each sample. Annotation area refers to the percentage of the image the user's annotation covers. There is no discernible difference between correct and incorrect classifications for time or area. 
    }
    \label{fig:time_area}
\end{figure}

\begin{figure}[!htb]
            \centering
          \includegraphics[width=0.38\columnwidth]{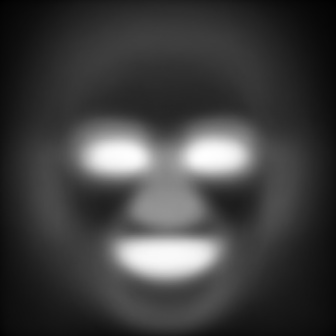}
          \includegraphics[width=0.38\columnwidth]{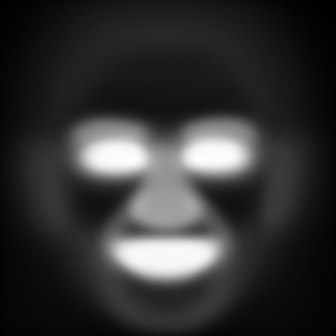}
    \caption{Average human annotation for both correct ({\bf left}) and incorrect ({\bf right}) annotations in the control phase.}
    \label{fig:annotations}
\end{figure}

Fig. \ref{fig:time_area} (left) summarizes the time spent per image in the control portion of the experiment. It is clear that there is {\bf no discernible difference between the time spent on a correct versus incorrect classification}. 
Similarly, as seen in Fig. \ref{fig:time_area} (right), there is {\bf no clear difference between annotation area for solving the task correctly versus incorrectly}. The most common annotation area is 10\% to 20\% of the image. Additionally, Fig. \ref{fig:annotations} shows the average human saliency annotation for both correctly and incorrectly classified samples. 
Note that all face images are aligned, so facial features should appear in consistent locations. 
Both correct and incorrect annotations have strong focus on the eyes, mouth and nose. The main difference is that the correct annotations appear less focused  on strong facial features.

\paragraph{Answering RQ4 (Accuracy for various types of ``hints'' from AI).}

\begin{figure}[]
    \centering
    \includegraphics[width=0.75\columnwidth]{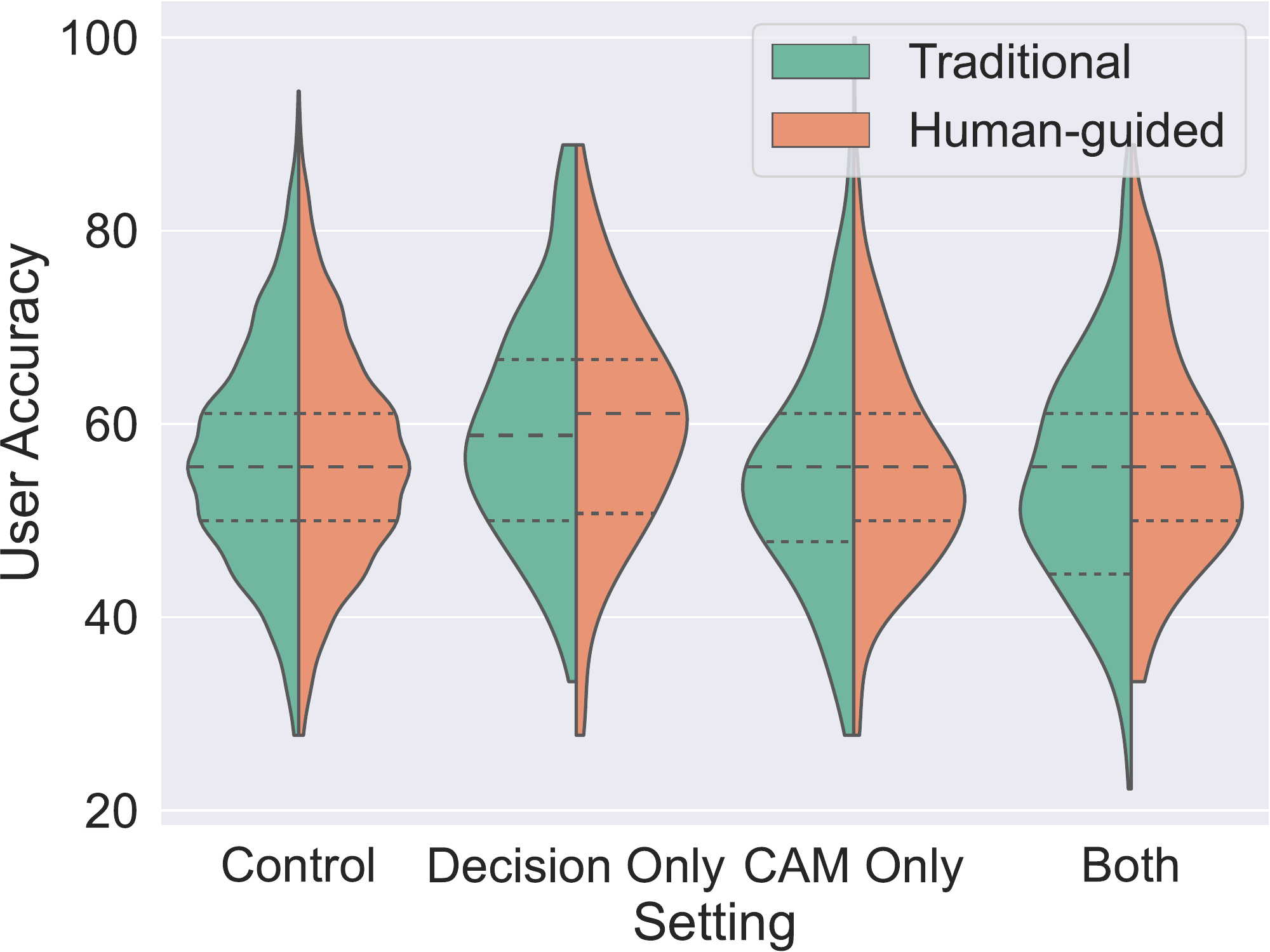}
    \caption{User accuracy for different types of the AI cues.}
    \label{fig:ai_setting_acc}
    \null
\end{figure}

Figure \ref{fig:ai_setting_acc} presents distributions of human accuracy for all four settings (no AI support, AI decision only, AI salience regions, both AI decision and salience regions). The first, surprising observation, is that presenting the AI salience, either alone or accompanied by the AI decision, didn't help humans  achieve higher accuracy (mean accuracy hovers around 56\%). Even more surprising is that this is true for both AI trained in a human-guided manner, and trained traditionally without human supervision. 

There is, however, a slight increase in accuracy when humans can benefit from the AI's decision (only), displayed for each image (59\%, compared to 56\% for all other scenarios). This accuracy further increases to 61\% when human-guided model is paired with humans. This increase is reasonable as the human-guided model is more accurate.

For traditionally trained models, the Mean Squared Error (MSE) between human annotations and the model saliency displayed in the CAM Only and CAM \& Decision settings is 0.24 and 0.23 respectively while the Structural Similarity (SSIM) is 0.13 and 0.12. Whereas, for the human guided models these values are improved to \textbf{0.13} and \textbf{0.12} for MSE and \textbf{0.14} and \textbf{0.14} for SSIM in the CAM Only and CAM \& Decision settings respectively.
Humans annotations tended to agree with the CAMs from the human-guided models more than traditionally trained models.

{\bf Answering RQ4}, {\bf humans did not benefit from seeing model salience, presented as class activation maps}, even if these maps closely resemble human saliency, but \textbf{do benefit from knowing the model's decision}. 
This may call for more effective ways of presenting model salience to humans.

\paragraph{Answering RQ5 (Agreement between AI and humans).}

\begin{figure}
    \centering
    \includegraphics[width=0.9\linewidth]{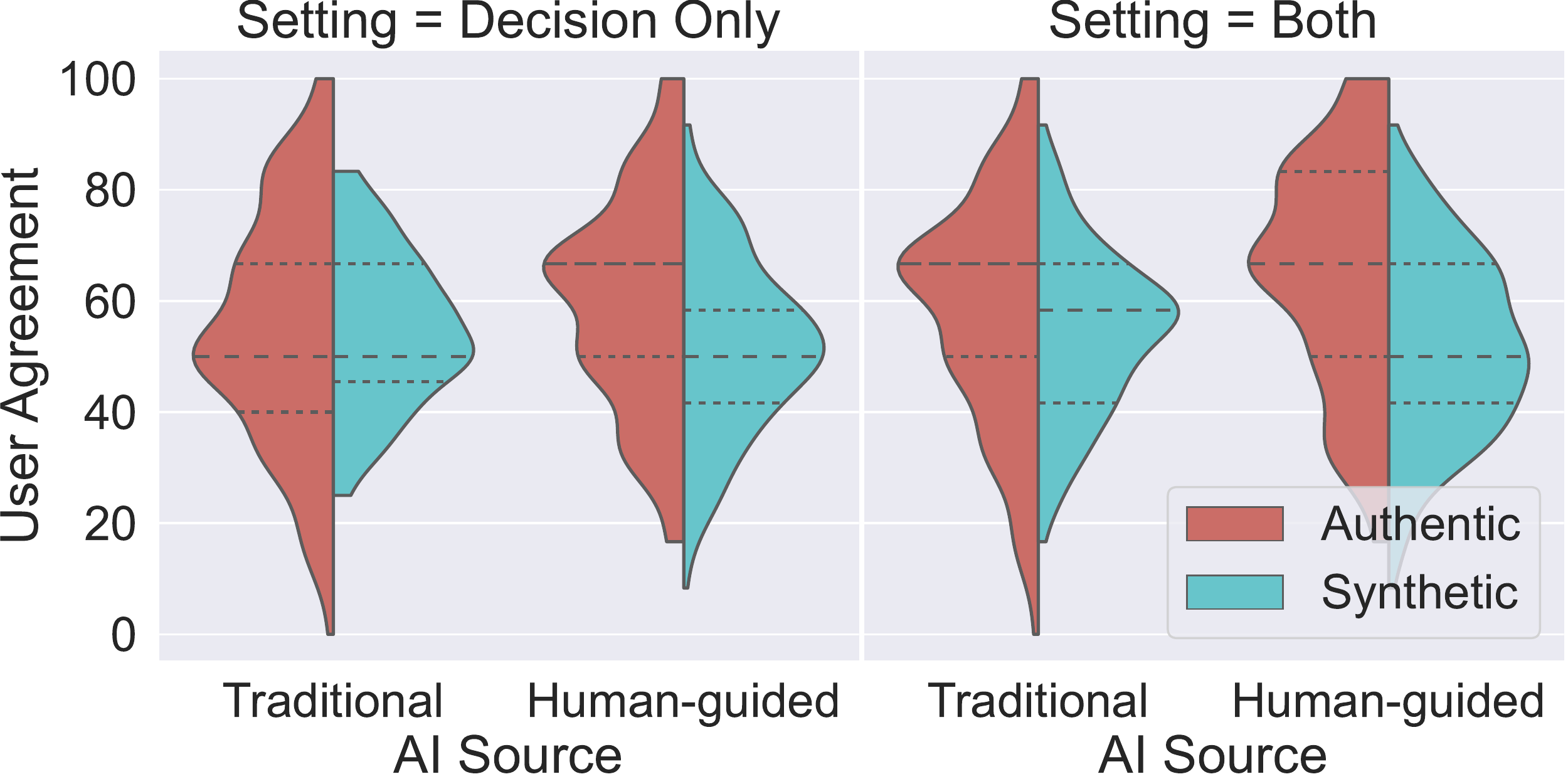}
    \caption{User agreement with the AI decision.}
    \label{fig:agreement}
    \null
\end{figure}

The agreement metric can be interpreted as humans trusting the model's decision.
Agreement of users for both traditionally trained and human-guided models can be seen in Fig. \ref{fig:agreement} for ``Decision Only'' and ``Both'' settings (the model's classification decision was not presented in the CAM Only setting). 
100\% means humans use the AI decision every time, and 0\% means they always chose the opposite of the AI. 
Agreement with AI is generally higher when both the model's CAM and decision are supplied to humans. Interestingly, referring back to Fig. \ref{fig:ai_setting_acc}, human accuracy is higher when only the AI decision is displayed compared to when decision and CAM are displayed. It appears that the knowledge of \textit{what} the AI thinks and \textit{why} leads people to be more confident in the AI decision, leading to poorer overall classification (due to AI's relatively low accuracy). 

Secondly, when comparing the agreement between humans and the two AI sources, human-guided models result in higher agreement than traditionally-trained models (60\% vs 50\%). Both human accuracy (as in Fig. \ref{fig:ai_setting_acc}) and agreement is higher for human-guided models, meaning humans recognize the increase in AI quality and trust the decisions more. 

For all models in all settings in Fig. \ref{fig:agreement}, humans have higher agreement with AI for authentic faces versus when AI predicts synthetic. This reflects the difficulty of the task of synthetic face detection. When referencing Fig. \ref{fig:gan_acc}, four of the six GAN sources are classified at less than 50\% accuracy. This means humans are more likely to classify samples as authentic than synthetic, and so when an AI model suggests authentic, it may reinforce their intuition and make them likely to agree. Conversely, when they think an image is authentic, but the AI suggests synthetic, they may choose to trust their judgement over AI, resulting in a lower agreement.
\paragraph{Answering RQ6 (Human's implicit perception of the AI's true accuracy from its decisions).}

Up to this point, it was shown that humans tended to have higher agreement with more accurate AI models (human-guided) compared to a slightly worse performing model. This raises the question: are humans inherently sensitive to AI performance? If humans see that the AI decisions seem to be right more often, are they more likely to increase agreement in the AI? 
To investigate this, two new AI models were created. These additional ``AI'' models have an overall accuracy of 90\%, and 10\%. Users are told these decisions are from an AI, however, they were just generated image labels to observe the desired ``AI'' performance.
The AI input supplied is decision only, as this was found to be the best way of portraying AI information according to Fig. \ref{fig:ai_setting_acc}.
For each AI accuracy setting, we collect data from 50 new subjects. This resulted in an additional 3,596 individual annotations. 
We can directly compare these new experiments to the human subjects that examined the same subset of images but with support of the original traditionally-trained and human-guided models.

While we are interested in the performance of humans when the AI accuracy goes up, we are also interested to see what happens when the AI gets much worse. Do they still trust these models? Or do they disregard them and just complete the task without the AI. The results of this experiment are summarized in Tab. \ref{tab:ai_changing_acc}. For the experiment when the AI was poor (10\% accuracy) Tab. \ref{tab:ai_changing_acc} shows that the human accuracy does not decrease significantly, staying above 50\%. For this experiment, the human/AI agreement drops, showing the lowest overall agreement. This suggests humans decided to ignore the AI input and make classification decisions independently, achieving average user accuracy similar to the control portion of the experiment.

From Tab. \ref{tab:ai_changing_acc} it can be seen that the setting in which humans performed best was when the AI was 90\% accurate. In this setting, both the overall accuracy and overall human/AI agreement was highest. Interestingly, even though the AI was 90\% accurate, the overall human accuracy is still significantly below that ($63.23\%\pm11.41\%$) suggesting that even though there is an increase in human performance when the AI improves, subjects are still skeptical and oftentimes trust their own decisions over that of the AI. {\bf Answering RQ6 we can state that humans tend to detect the performance of the underlying AI, even without any accuracy information being explicitly given}.

\begin{table*}[]
\centering
\caption{Human accuracy and human-AI agreement depending on the AI accuracy supporting examination. Humans {\bf were not} informed about the AI's accuracy, yet they could make their own assessment of the AI's reliability by observing its decisions.}
\footnotesize
\begin{tabular}{|c|c|c|c|c|}
\hline
\textbf{Actual AI Accuracy:} & \textbf{Traditional (50\%)} & \textbf{Human-Guided (60\%)} & \textbf{90\%} & \textbf{10\%} \\ \hline\hline
\textbf{User Accuracy}          & 57.54$\pm$16.6  & 60.44$\pm$15.58 & \textbf{63.23$\pm$11.41} & 56.61$\pm$11.12          \\ \hline
\textbf{User Acc. on Authentic} & 66.15$\pm$24.86 & 63.01$\pm$25.28 & \textbf{73.2$\pm$21.29}  & 61.33$\pm$24.14          \\ \hline
\textbf{User Acc. on Synthetic} & 55.11$\pm$16.62 & 57.14$\pm$15.75 & \textbf{58.24$\pm$15.16} & \textit{54.27$\pm$15.52} \\ \hline
\textbf{Human/AI Agreement}     & 57.3$\pm$15.54  & 55.83$\pm$18.62 & \textbf{62.78$\pm$10.91} & \textit{47.61$\pm$11.2}  \\ \hline
\end{tabular}
\label{tab:ai_changing_acc}
\end{table*}

\begin{table*}[ht!]
\centering
\caption{Same as in Table \ref{tab:ai_changing_acc}, except that humans {\bf were} informed about the AI's accuracy.}
\footnotesize
\begin{tabular}{|c|cc|cc|}
\hline
\textbf{Actual AI Accuracy:} & \multicolumn{2}{c|}{\textbf{60\%}} & \multicolumn{2}{c|}{\textbf{90\%}} \\ \hline
\textbf{Reported AI Accuracy}   & \multicolumn{1}{c|}{\textbf{60\%}} & \textbf{90\%} & \multicolumn{1}{c|}{\textbf{60\%}}      & \textbf{90\%}       \\ \hline\hline
\textbf{User Accuracy}          & \multicolumn{1}{c|}{59.56$\pm$11.96}    & 60.81$\pm$11.98    & \multicolumn{1}{c|}{62.36$\pm$10.37}         & \textbf{65.15$\pm$11.74} \\ \hline
\textbf{User Acc. on Authentic} & \multicolumn{1}{c|}{68.27$\pm$22.38}    & 65.2$\pm$23.07     & \multicolumn{1}{c|}{\textbf{70.0$\pm$24.12}} & 66.07$\pm$20.45          \\ \hline
\textbf{User Acc. on Synthetic} & \multicolumn{1}{c|}{55.24$\pm$18.35}    & 58.64$\pm$16.7     & \multicolumn{1}{c|}{58.56$\pm$15.11}         & \textbf{64.67$\pm$15.24} \\ \hline
\textbf{Human/AI Agreement}     & \multicolumn{1}{c|}{55.69$\pm$12.92}    & 59.59$\pm$16.71    & \multicolumn{1}{c|}{62.37$\pm$10.63}         & \textbf{65.38$\pm$12.04} \\ \hline
\end{tabular}
\label{tab:ai_reported_acc}
\end{table*}

\paragraph{Answering RQ7 (What builds trust in AI?).}

Seeing that humans implicitly detect changing AI performance, placing more trust in seemingly better models, does human performance change when they are explicitly aware of the underlying AI performance? In all previous experiments, the human subjects were unaware of what the underlying AI was and how well it performed. For this final experiment, the tool was slightly modified to add the following sentence above the image: ``This AI generates a correct decision X\% of the time.'' with X = 60\% and X = 90\%. As previously, the AI input supplied is decision only.

We investigate whether supplying this information has an effect for two AI modes: using the 60\% accurate human-guided decisions and using the same 90\% accurate model from RQ6. 
For each of these settings, we run two experiments. Firstly, truthful information about the AI accuracy is supplied \ie ``This AI generates a correct decision 60\% or 90\% of the time.'' for the respective models. Secondly, false information about the AI is supplied \ie when using the 90\% accurate AI it says ``This AI generates a correct decision 60\% of the time.'' and vice versa. Data is collected from 50 subjects for each of these four settings on the exact same images as for RQ6, resulting in 7,187 new annotations. 

Tab. \ref{tab:ai_reported_acc} details the results of this experiment. Similar to the previous subsection, the best results were achieved when the AI performance was 90\%. The performance of humans also increased when they were told the AI's accuracy, compared to the scenario when this accuracy had to be judged by humans only by observing the AI's performance (cf. Tab. \ref{tab:ai_changing_acc}). When the model is 60\% accurate untruthfully telling them it is 90\% accurate resulted in an increase in agreement, suggesting they trusted it slightly more. Similarly, when users are told the 90\% accurate model is just 60\% accurate, agreement decreases compared to when humans are truthfully told it is 90\% accurate, suggesting this information is factored into the decision-making process.

Interestingly, the agreement for humans is higher for the 90\% accurate model, even when they are told the AI accuracy is 60\%. Similarly to when no AI information is given, humans possess the capability of recognizing the model performance and making judgements on whether to believe it or not, independently of what is reported to them. 
Still, when the model is 90\% accurate and users are truthfully informed of this, human accuracy is ~65\%. This result shows humans don't blindly place trust in these models, often ignoring the AI to make their own judgement. 
\textbf{Answering RQ7, humans place more trust in AI support when given information about the underlying model performance.}

\section{Conclusions}

The ability of humans to detect whether a face image is authentic or synthetic is essential in varied social media and forensic scenarios. The latest GAN-created synthetic face images make this task extremely difficult for humans, and the best algorithmic approaches fare only slightly better. This paper explores how human performance on this task might be improved given access to hints from a better-performing algorithmic approach.  

We found that humans correctly classify authentic face images almost two thirds of the time. Correctly classifying synthetic images is more difficult, where human accuracy is near random, at about 52\%. There is no clear indication that human accuracy is associated with either the amount of time spent considering an image or the amount of the image that is annotated as salient to the decision.

One main result is that human accuracy is improved when they are shown the algorithm’s classification result. However, showing the model's saliency, either alone without the algorithm’s classification result, or in addition to the algorithm’s classification result, is not as helpful. 

We found huge differences in the quality of GAN-produced face images in the context of this task. At one end, ProGAN and StarGANv2 images are relatively easily detected as synthetic, at 94\% and 90\% accuracy, respectively. At the other end, StyleGAN2 and StyleGAN2-ADA images were correctly detected as synthetic only 25\% and 26\% of the time, respectively. The curious element of this result is that the best GANs produce images more convincing as authentic than actual authentic images.

We demonstrate that humans implicitly recognize when the underlying AI performance changes, adjusting their trust in these models accordingly. Additionally, giving explicit information about the performance of the AI improves human performance and trust in the models. This paper provides an insight into how synthetic image detection technology would be adopted by humans upon deployment.

\section*{Acknowledgments}
Research presented in this paper was supported by the U.S. Department of Defense (Contract No. W52P1J2093009). The views and conclusions contained in this paper are those of the authors and should not be interpreted as representing the official policies, either expressed or implied, of the U.S. Department of Defense or the U.S. Government. The U.S. Government is authorized to reproduce and distribute reprints for Government purposes, notwithstanding any copyright notation here on.

\bibliography{aaai23}

\begin{thebibliography}{25}
\providecommand{\natexlab}[1]{#1}

\bibitem[{Albiero et~al.(2021)Albiero, Chen, Yin, Pang, and
  Hassner}]{Albiero_CVPR_2021}
Albiero, V.; Chen, X.; Yin, X.; Pang, G.; and Hassner, T. 2021.
\newblock {img2pose: Face Alignment and Detection via 6DoF, Face Pose
  Estimation}.
\newblock In \emph{IEEE/CVF Conf. Comput. Vis. Pattern Recog. (CVPR)},
  7613--7623.

\bibitem[{Bansal et~al.(2021)Bansal, Wu, Zhou, Fok, Nushi, Kamar, Ribeiro, and
  Weld}]{Bansal_CHI_2021}
Bansal, G.; Wu, T.~S.; Zhou, J.; Fok, R.; Nushi, B.; Kamar, E.; Ribeiro, M.~T.;
  and Weld, D.~S. 2021.
\newblock {Does the Whole Exceed its Parts? The Effect of AI Explanations on
  Complementary Team Performance}.
\newblock In \emph{Proceedings of the 2021 CHI Conference on Human Factors in
  Computing Systems (CHI'21)}.

\bibitem[{Boyd, Bowyer, and Czajka(2022)}]{boyd2021human}
Boyd, A.; Bowyer, K.~W.; and Czajka, A. 2022.
\newblock Human-aided saliency maps improve generalization of deep learning.
\newblock In \emph{Proceedings of the IEEE/CVF Winter Conference on
  Applications of Computer Vision}, 2735--2744.

\bibitem[{Boyd et~al.(2021)Boyd, Tinsley, Bowyer, and Czajka}]{boyd2021cyborg}
Boyd, A.; Tinsley, P.; Bowyer, K.; and Czajka, A. 2021.
\newblock CYBORG: Blending Human Saliency Into the Loss Improves Deep Learning.
\newblock \emph{arXiv preprint arXiv:2112.00686}.

\bibitem[{Brock, Donahue, and Simonyan(2018)}]{brock2018large}
Brock, A.; Donahue, J.; and Simonyan, K. 2018.
\newblock {Large scale GAN training for high fidelity natural image synthesis}.
\newblock \emph{arXiv preprint arXiv:1809.11096}.

\bibitem[{Choi et~al.(2020)Choi, Uh, Yoo, and Ha}]{choi2020stargan}
Choi, Y.; Uh, Y.; Yoo, J.; and Ha, J.-W. 2020.
\newblock {StarGAN v2: Diverse image synthesis for multiple domains}.
\newblock In \emph{IEEE/CVF Conf. Comput. Vis. Pattern Recog. (CVPR)},
  8185--8194.

\bibitem[{Goodfellow et~al.(2014)Goodfellow, Pouget-Abadie, Mirza, Xu,
  Warde-Farley, Ozair, Courville, and Bengio}]{goodfellow2014generative}
Goodfellow, I.; Pouget-Abadie, J.; Mirza, M.; Xu, B.; Warde-Farley, D.; Ozair,
  S.; Courville, A.; and Bengio, Y. 2014.
\newblock Generative adversarial nets.
\newblock \emph{Advances in Neural Information Processing Systems}, 27.

\bibitem[{{He} et~al.(2019){He}, {Tavakoli}, {Borji}, and
  {Pugeault}}]{He_ICCV_2019}
{He}, S.; {Tavakoli}, H.~R.; {Borji}, A.; and {Pugeault}, N. 2019.
\newblock Human Attention in Image Captioning: Dataset and Analysis.
\newblock In \emph{2019 IEEE/CVF International Conference on Computer Vision
  (ICCV)}, 8528--8537.

\bibitem[{Jacobs et~al.(2021)Jacobs, Pradier, McCoy, Perlis, Doshi-Velez, and
  Gajos}]{Jacobs_Nature_2021}
Jacobs, M.; Pradier, M.~F.; McCoy, T.~H., Jr; Perlis, R.~H.; Doshi-Velez, F.;
  and Gajos, K.~Z. 2021.
\newblock How machine-learning recommendations influence clinician treatment
  selections: the example of the antidepressant selection.
\newblock \emph{Transl. Psychiatry}, 11(1): 108.

\bibitem[{Jain, Deb, and Engelsma(2022)}]{Jain_TBIOM_2022}
Jain, A.~K.; Deb, D.; and Engelsma, J.~J. 2022.
\newblock Biometrics: Trust, But Verify.
\newblock \emph{IEEE Transactions on Biometrics, Behavior, and Identity
  Science}, 4(3): 303--323.

\bibitem[{Karras et~al.(2017)Karras, Aila, Laine, and
  Lehtinen}]{karras2017progressive}
Karras, T.; Aila, T.; Laine, S.; and Lehtinen, J. 2017.
\newblock {Progressive Growing of GANs for Improved Quality, Stability, and
  Variation}.
\newblock \emph{arXiv preprint arXiv:1710.10196}.

\bibitem[{Karras et~al.(2020{\natexlab{a}})Karras, Aittala, Hellsten, Laine,
  Lehtinen, and Aila}]{Karras2020ada}
Karras, T.; Aittala, M.; Hellsten, J.; Laine, S.; Lehtinen, J.; and Aila, T.
  2020{\natexlab{a}}.
\newblock {Training Generative Adversarial Networks with Limited Data}.
\newblock In \emph{Adv. Neural Inform. Process. Syst. (NeurIPS)}.

\bibitem[{Karras et~al.(2021)Karras, Aittala, Laine, H\"ark\"onen, Hellsten,
  Lehtinen, and Aila}]{Karras2021}
Karras, T.; Aittala, M.; Laine, S.; H\"ark\"onen, E.; Hellsten, J.; Lehtinen,
  J.; and Aila, T. 2021.
\newblock {Alias-Free Generative Adversarial Networks}.
\newblock In \emph{Adv. Neural Inform. Process. Syst. (NeurIPS)}.

\bibitem[{Karras, Laine, and Aila(2019)}]{karras2019style}
Karras, T.; Laine, S.; and Aila, T. 2019.
\newblock {A Style-Based Generator Architecture for Generative Adversarial
  Networks}.
\newblock In \emph{IEEE/CVF Conf. Comput. Vis. Pattern Recog. (CVPR)},
  4401--4410.

\bibitem[{Karras et~al.(2020{\natexlab{b}})Karras, Laine, Aittala, Hellsten,
  Lehtinen, and Aila}]{karras2020analyzing}
Karras, T.; Laine, S.; Aittala, M.; Hellsten, J.; Lehtinen, J.; and Aila, T.
  2020{\natexlab{b}}.
\newblock {Analyzing and Improving the Image Quality of StyleGAN}.
\newblock In \emph{IEEE/CVF Conf. Comput. Vis. Pattern Recog. (CVPR)},
  8110--8119.

\bibitem[{Liefooghe et~al.(2022)Liefooghe, Oliveira, Leisten, Hoogers, Aarts,
  and Hortensius}]{Liefooghe_PsyArXiv_2022}
Liefooghe, B.; Oliveira, M. J.~B.; Leisten, L.~M.; Hoogers, E.; Aarts, H.; and
  Hortensius, R. 2022.
\newblock Faces merely labelled as artificial are trusted less.

\bibitem[{Moreira et~al.(2019)Moreira, Trokielewicz, Czajka, Bowyer, and
  Flynn}]{Moreira_WACV_2019}
Moreira, D.; Trokielewicz, M.; Czajka, A.; Bowyer, K.; and Flynn, P. 2019.
\newblock Performance of Humans in Iris Recognition: The Impact of Iris
  Condition and Annotation-Driven Verification.
\newblock In \emph{2019 IEEE Winter Conference on Applications of Computer
  Vision (WACV)}, 941--949.

\bibitem[{Nakano and Yamamoto(2022)}]{Nakano_HSSC_2022}
Nakano, T.; and Yamamoto, T. 2022.
\newblock You trust a face like yours.
\newblock \emph{Humanities and Social Sciences Communications}, 9(1).

\bibitem[{Nightingale and Farid(2022)}]{doi:10.1073/pnas.2120481119}
Nightingale, S.~J.; and Farid, H. 2022.
\newblock AI-synthesized faces are indistinguishable from real faces and more
  trustworthy.
\newblock \emph{Proceedings of the National Academy of Sciences}, 119(8):
  e2120481119.

\bibitem[{{O'Toole} et~al.(2007){O'Toole}, {Abdi}, {Jiang}, and
  {Phillips}}]{OToole_TSMC_2007}
{O'Toole}, A.~J.; {Abdi}, H.; {Jiang}, F.; and {Phillips}, P.~J. 2007.
\newblock Fusing Face-Verification Algorithms and Humans.
\newblock \emph{IEEE Transactions on Systems, Man, and Cybernetics, Part B
  (Cybernetics)}, 37(5): 1149--1155.

\bibitem[{Shen et~al.(2021)Shen, RichardWebster, O'Toole, Bowyer, and
  Scheirer}]{shen2021study}
Shen, B.; RichardWebster, B.; O'Toole, A.; Bowyer, K.; and Scheirer, W.~J.
  2021.
\newblock A Study of the Human Perception of Synthetic Faces.
\newblock \emph{arXiv preprint arXiv:2111.04230}.

\bibitem[{Trokielewicz, Czajka, and Maciejewicz(2019)}]{Trokielewicz_BTAS_2019}
Trokielewicz, M.; Czajka, A.; and Maciejewicz, P. 2019.
\newblock {Perception of Image Features in Post-Mortem Iris Recognition: Humans
  vs Machines}.
\newblock In \emph{2019 IEEE 10th International Conference on Biometrics
  Theory, Applications and Systems (BTAS), September 23-26, 2019, Tampa, FL,
  USA}, 1--9. Tampa, FL, USA: IEEE.

\bibitem[{Tucciarelli, Vehar, and Tsakiris(2020)}]{Tucciarelli_PsyArXiv_2020}
Tucciarelli, R.; Vehar, N.; and Tsakiris, M. 2020.
\newblock On the realness of people who do not exist: the social processing of
  artificial faces.

\bibitem[{Wong, Wang, and Hryniowski(2020)}]{wong2020much}
Wong, A.; Wang, X.~Y.; and Hryniowski, A. 2020.
\newblock How much can we really trust you? towards simple, interpretable trust
  quantification metrics for deep neural networks.
\newblock \emph{arXiv preprint arXiv:2009.05835}.

\bibitem[{Zhou et~al.(2016)Zhou, Khosla, Lapedriza, Oliva, and
  Torralba}]{zhou2016learning}
Zhou, B.; Khosla, A.; Lapedriza, A.; Oliva, A.; and Torralba, A. 2016.
\newblock Learning deep features for discriminative localization.
\newblock In \emph{IEEE/CVF Conf. Comput. Vis. Pattern Recog. (CVPR)},
  2921--2929.

\end{thebibliography}

\appendix

\section{Additional Examples}

\begin{figure*}[!htb]
    \centering
    \includegraphics[width=\linewidth]{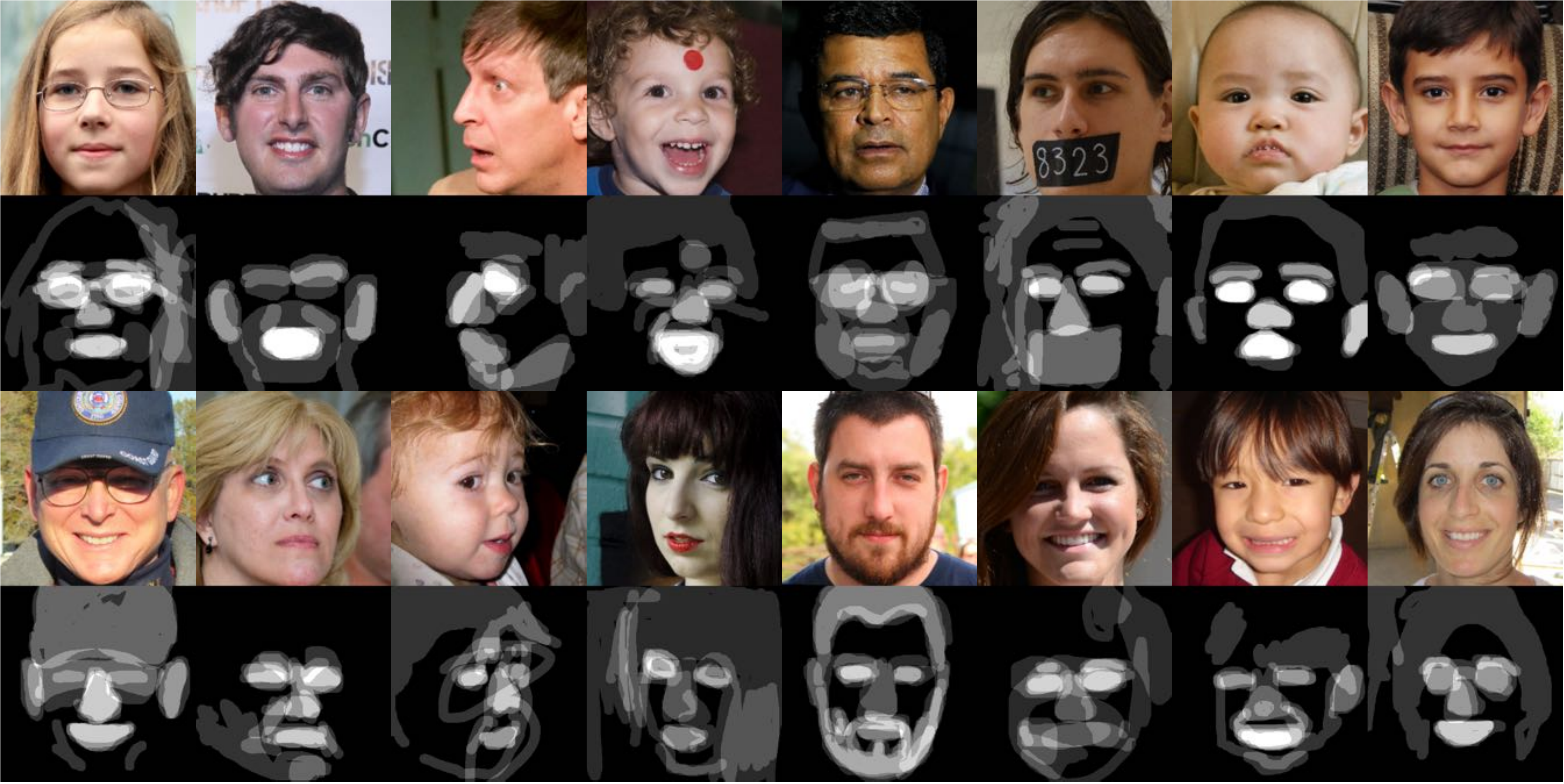}
    \caption{Sixteen random examples of \textbf{authentic} samples \cite{karras2019style} (1st \& 3rd row) and corresponding human annotations (2nd \& 4th row) from the control phase of the collection.
    }
    \label{fig:ffhq}
\end{figure*}

\begin{figure*}[!ht]
    \centering
    \includegraphics[width=\linewidth]{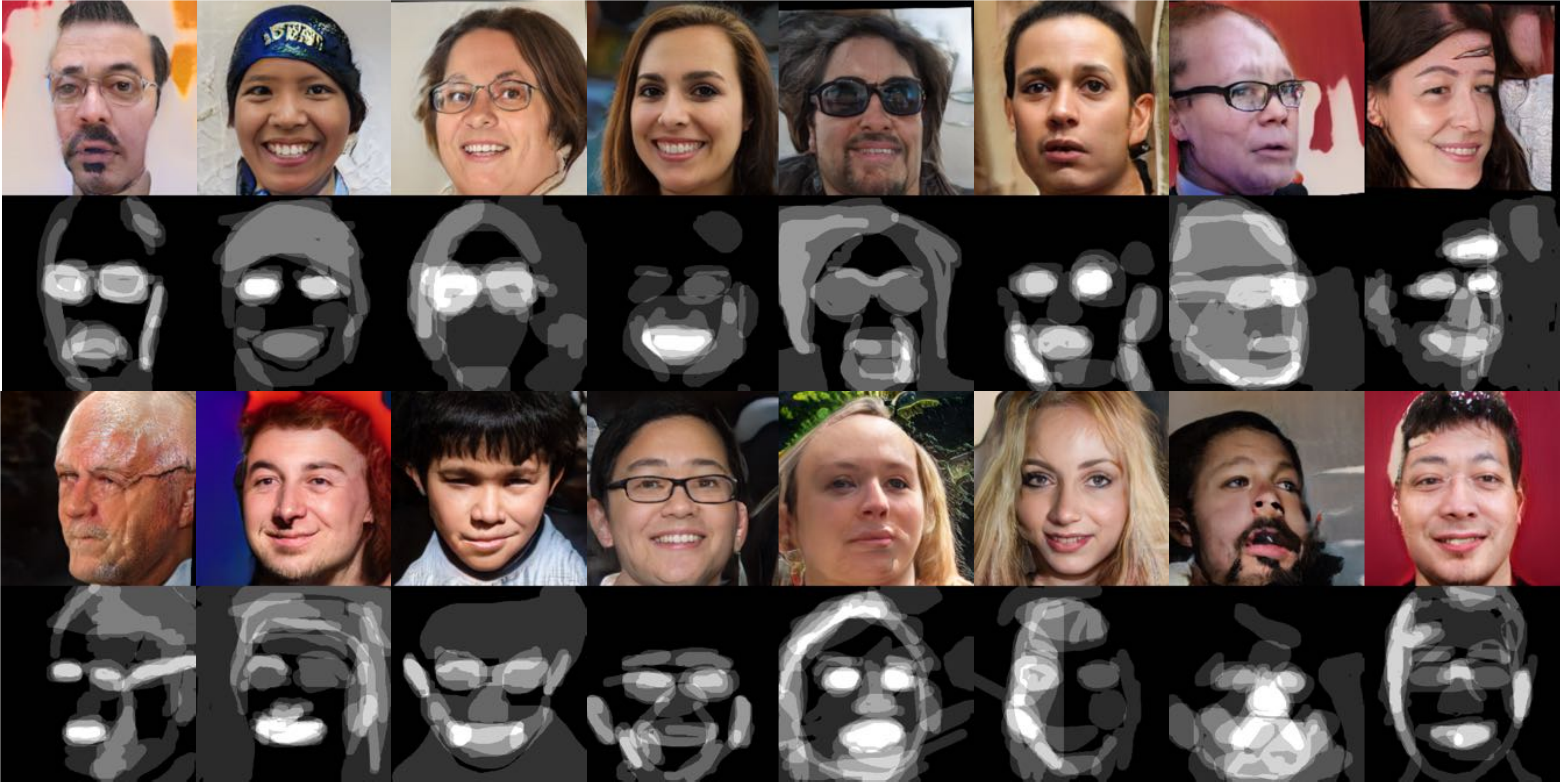}
    \caption{Sixteen random examples of \textbf{ProGAN} \cite{karras2017progressive} samples (1st \& 3rd row) and corresponding human annotations (2nd \& 4th row) from the control phase of the collection.
    }
    \label{fig:progan}
\end{figure*}

\begin{figure*}[!ht]
    \centering
    \includegraphics[width=\linewidth]{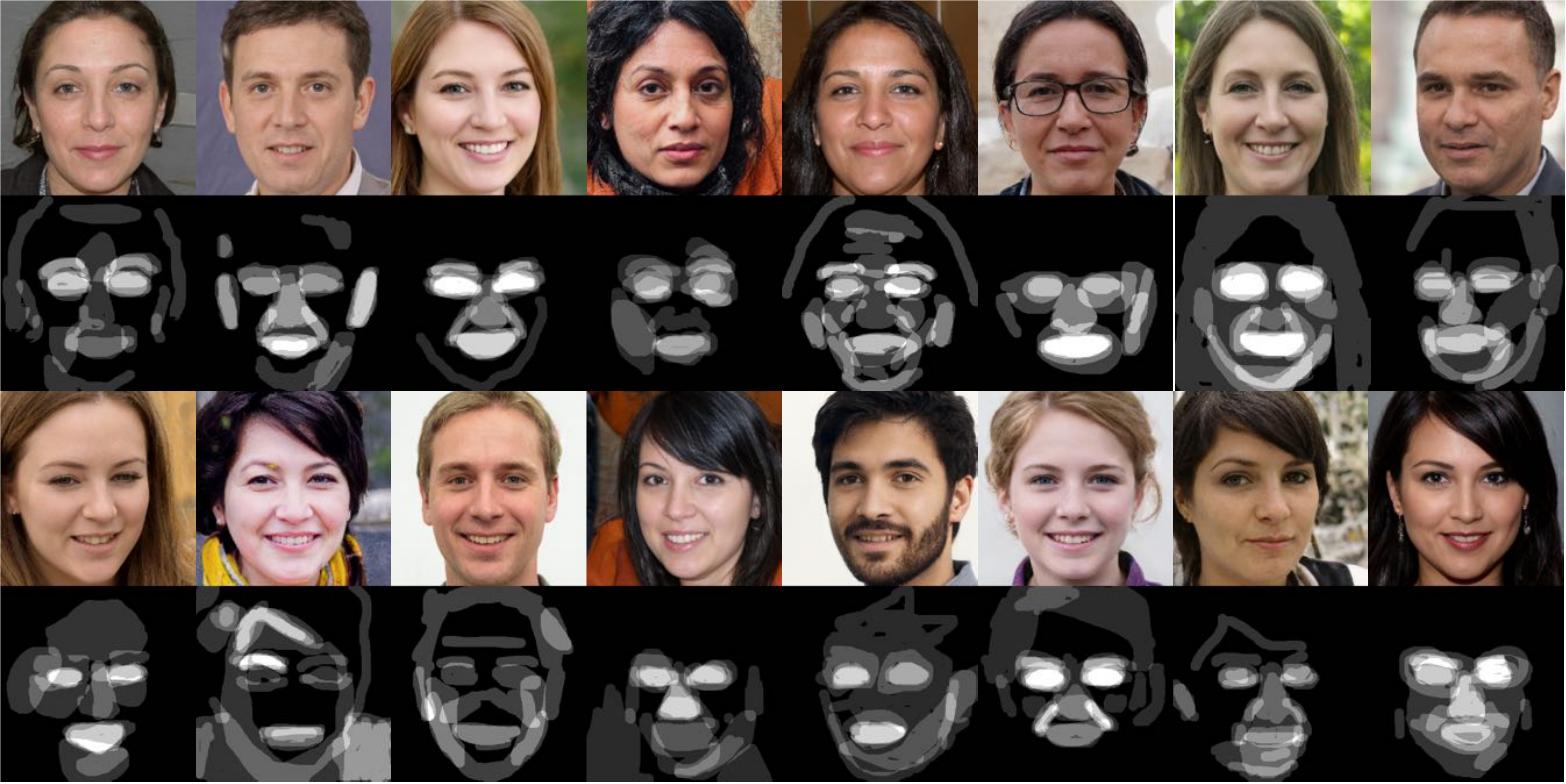}
    \caption{Sixteen random examples of \textbf{StyleGAN} \cite{karras2019style} samples (1st \& 3rd row) and corresponding human annotations (2nd \& 4th row) from the control phase of the collection.
    }
    \label{fig:sg1}
\end{figure*}

\begin{figure*}[!ht]
    \centering
    \includegraphics[width=\linewidth]{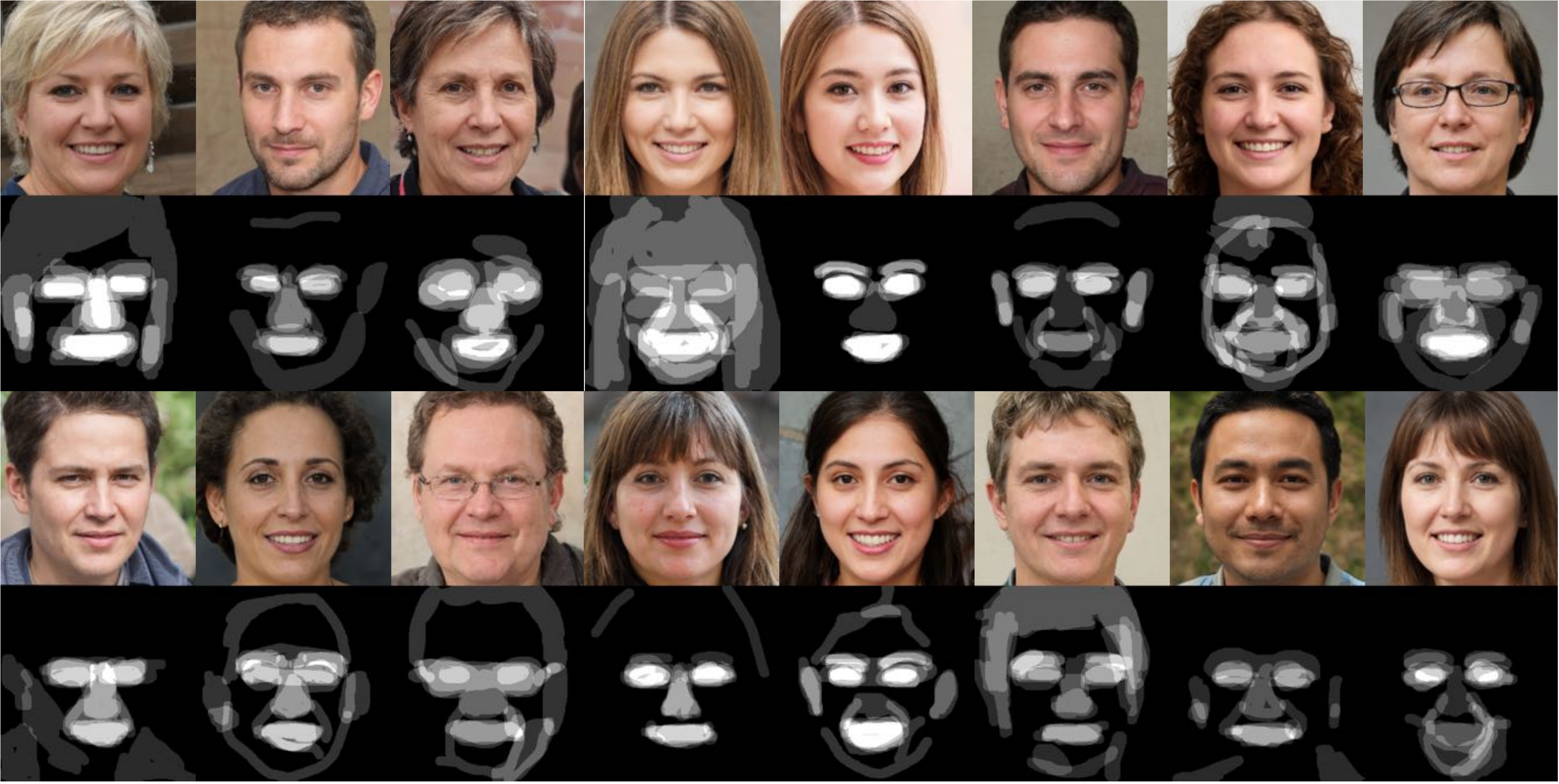}
    \caption{Sixteen random examples of \textbf{StyleGAN2} \cite{karras2020analyzing} samples (1st \& 3rd row) and corresponding human annotations (2nd \& 4th row) from the control phase of the collection.
    }
    \label{fig:sg2}
\end{figure*}

\begin{figure*}[!ht]
    \centering
    \includegraphics[width=\linewidth]{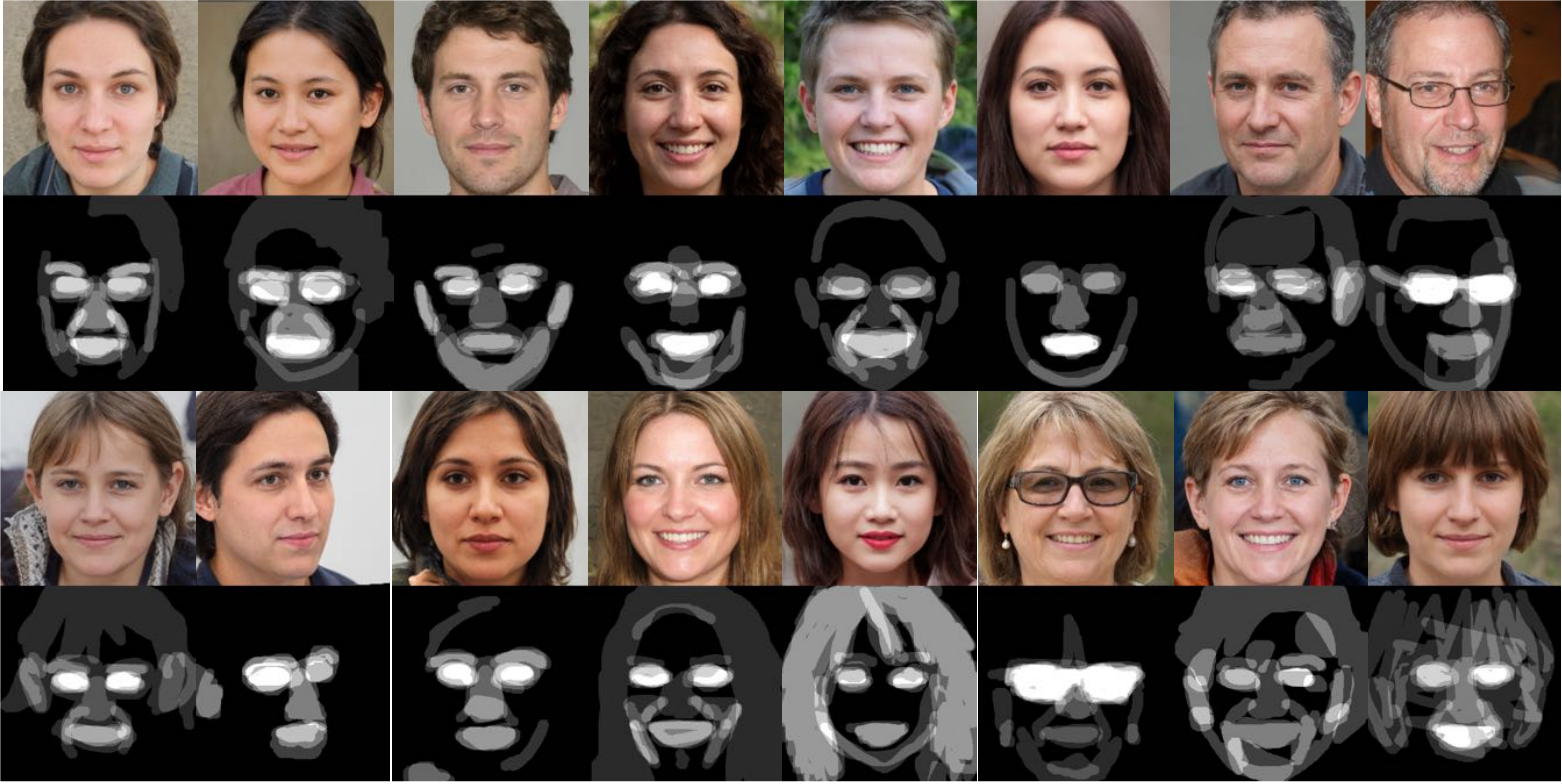}
    \caption{Sixteen random examples of \textbf{StyleGAN2-ADA} \cite{Karras2020ada} samples (1st \& 3rd row) and corresponding human annotations (2nd \& 4th row) from the control phase of the collection.
    }
    \label{fig:sg2ada}
\end{figure*}

\begin{figure*}[!ht]
    \centering
    \includegraphics[width=\linewidth]{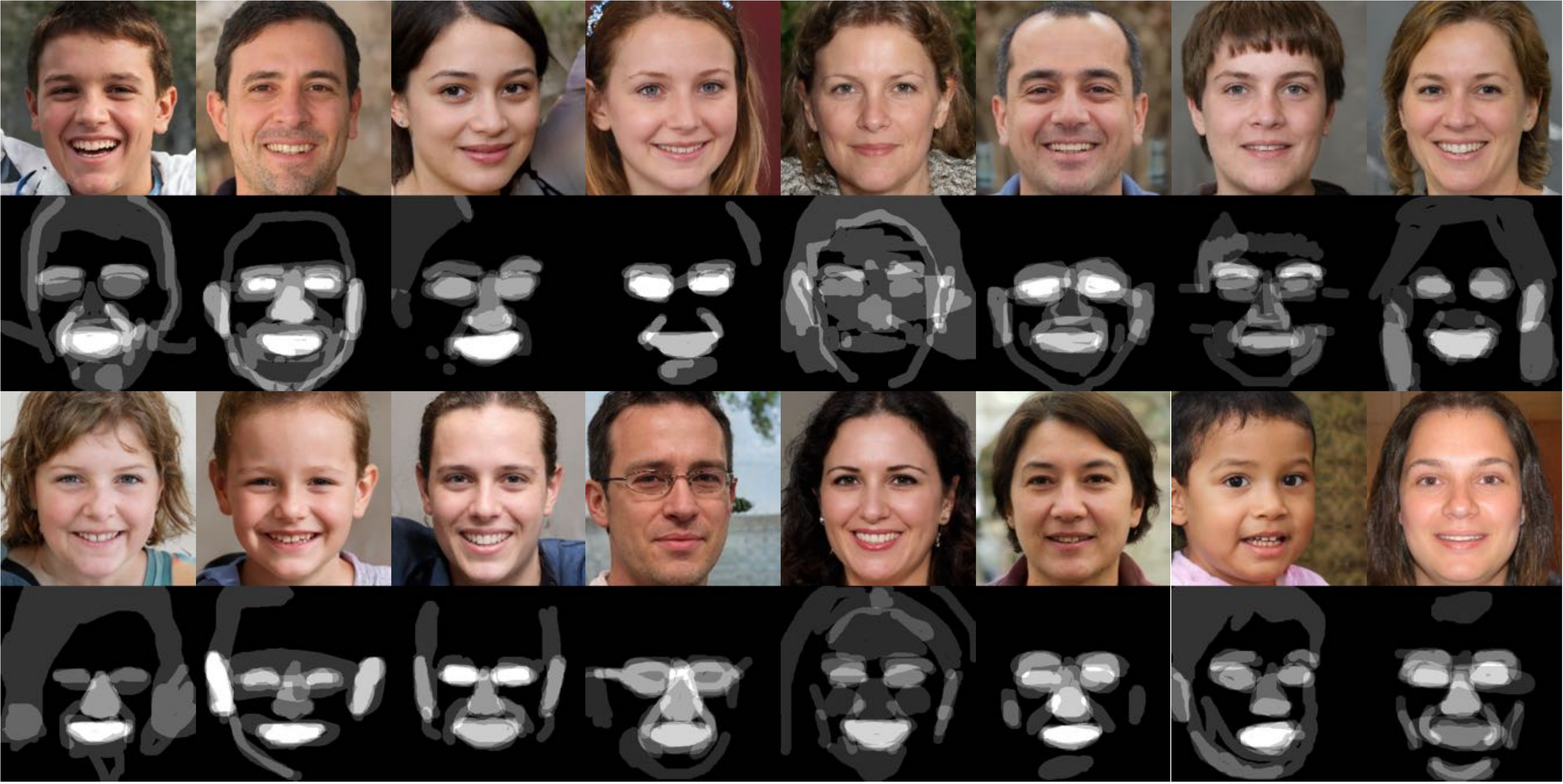}
    \caption{Sixteen random examples of \textbf{StyleGAN3} \cite{Karras2021} samples (1st \& 3rd row) and corresponding human annotations (2nd \& 4th row) from the control phase of the collection.
    }
    \label{fig:sg3}
\end{figure*}

\begin{figure*}[!ht]
    \centering
    \includegraphics[width=\linewidth]{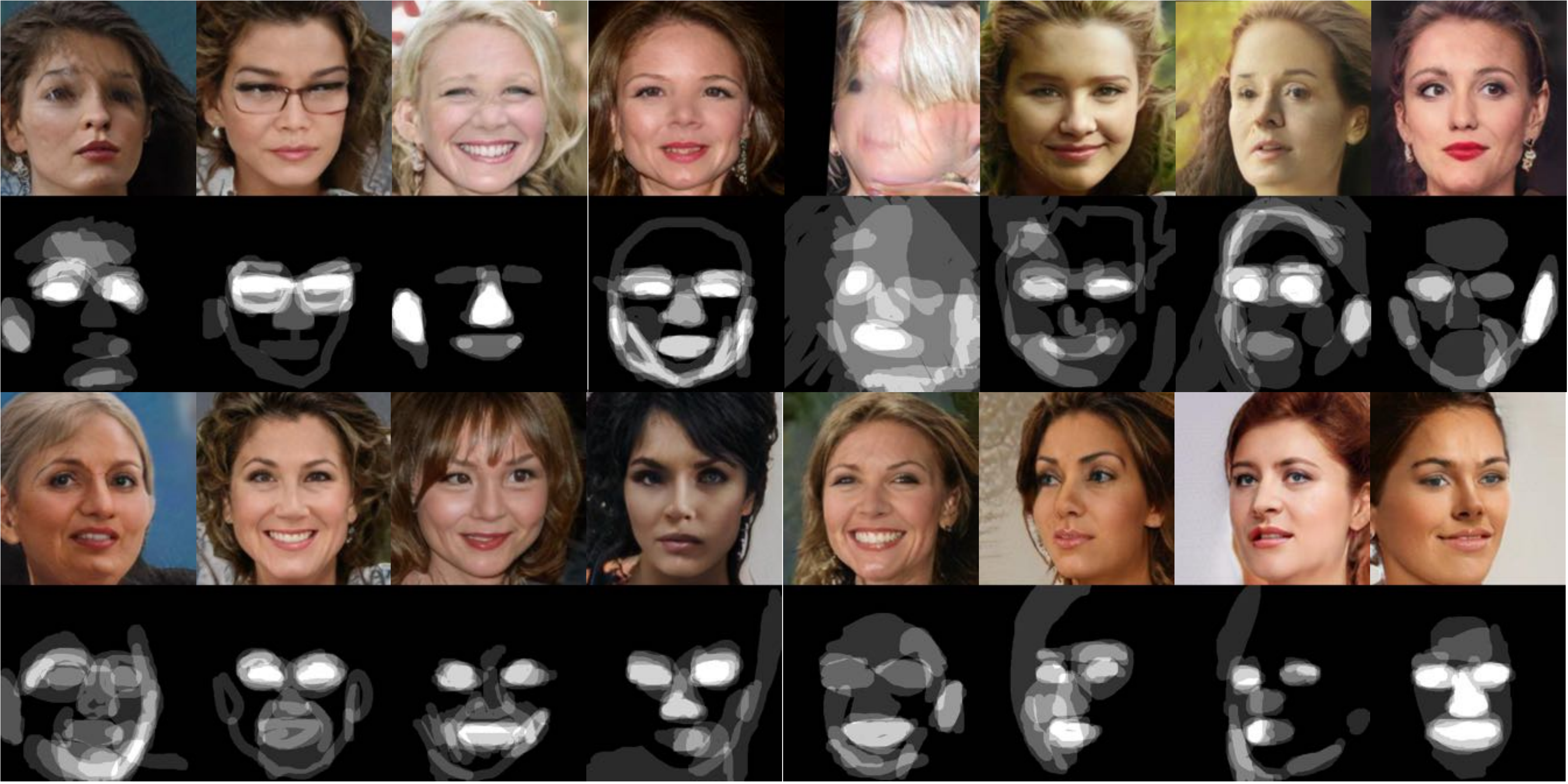}
    \caption{Sixteen random examples of \textbf{StarGANv2} \cite{choi2020stargan} samples (1st \& 3rd row) and corresponding human annotations (2nd \& 4th row) from the control phase of the collection.
    }
    \label{fig:stargan}
\end{figure*}

To better demonstrate the difference between the various image sources containing one authentic and six different GAN synthesizers, sixteen random samples are shown from each one (Fig. \ref{fig:ffhq}-Fig. \ref{fig:stargan} on following pages). Additionally, the combination of all human annotations on each image from the control phase of the collection is shown below each individual sample. 

\section{Source Code}

The code for the online tool and the statistics generation scripts will be released with the published version of this work, along with the datasets, to allow for reproducibility of this work.

\end{document}